\documentclass[conference, a4paper]{IEEEtran}
\usepackage{times}
\usepackage[marginal]{footmisc}
\usepackage[]{hyperref}
\usepackage{listings}
\usepackage{soul}
\usepackage{xcolor}
\usepackage{doi}
\usepackage{setspace}

\usepackage{amsmath}
\usepackage{verbatim}
\usepackage{booktabs}
\usepackage{bbding}

\lstdefinestyle{Python}{
    language        =   Python, 
    basicstyle      =   \small,
    numberstyle     =   \small,
    keywordstyle    =   \color{blue},
    keywordstyle    =   [2] \color{teal},
    stringstyle     =   \color{magenta},
    commentstyle    =   \color{red}\ttfamily,
    breaklines      =   true,   
    columns         =   fixed,  
    basewidth       =   0.5em,
}

\hypersetup{colorlinks = true, allcolors = blue}
\usepackage{amssymb}
\usepackage[T1]{fontenc}
\usepackage{titlesec}
\usepackage[numbers,sort,square]{natbib}
\titlespacing*{\section}{0pt}{0.1\baselineskip}{0.1\baselineskip}
\titlespacing*{\subsection}{0pt}{0.1\baselineskip}{0.1\baselineskip}

\usepackage[ruled,linesnumbered]{algorithm2e}
\usepackage{graphicx}

\IEEEoverridecommandlockouts
\IEEEpubid{\makebox[\columnwidth]{\begin{minipage}{\textwidth}\ \\[12pt] \centering DOI reference number: 10.18293/SEKE23-110 \end{minipage}} \hspace{\columnsep}\makebox[\columnwidth]{ }}
\begin{document}
\title{Using Z3 for Formal Modeling and Verification of FNN Global Robustness}
\author{\IEEEauthorblockN{Yihao Zhang\IEEEauthorrefmark{1}, Zeming Wei\IEEEauthorrefmark{1}, Xiyue Zhang\IEEEauthorrefmark{1}\IEEEauthorrefmark{2}\thanks{$^\dag$ Current Address: Department of Computer Science, University of Oxford, Oxford, UK.}, Meng Sun\thanks{$^\ddag$ Corresponding author.}\IEEEauthorrefmark{1}\IEEEauthorrefmark{3}}

\IEEEauthorblockA{\IEEEauthorrefmark{1}School of Mathematical Sciences, Peking University, Beijing, China\\
\{zhangyihao, weizeming\}@stu.pku.edu.cn, \{zhangxiyue,sunm\}@pku.edu.cn}

}

\maketitle       
\begin{abstract}
\small 
While Feedforward Neural Networks (FNNs) have achieved remarkable success in various tasks, they are vulnerable to adversarial examples. 
Several techniques have been developed to verify the adversarial robustness of FNNs, 
but most of them focus on  robustness verification against the local perturbation neighborhood of a single data point.
There is still a large research gap in global robustness analysis. 
The global-robustness verifiable framework DeepGlobal has been proposed to identify \textit{all} possible Adversarial Dangerous Regions (ADRs) of FNNs, 
not limited to data samples in a test set. 
In this paper, we propose a complete 
specification and implementation of DeepGlobal utilizing the SMT solver Z3 for more explicit definition, and
propose 
several improvements to DeepGlobal for
more efficient verification. 
To evaluate the effectiveness of our  
implementation and improvements, 
we conduct extensive experiments on a set of benchmark datasets. 
Visualization of our experiment results shows the validity and effectiveness of the approach.

\end{abstract}
\section{Introduction}
Feedforward Neural Networks (FNNs) have achieved remarkable success in various fields. Despite their success, the existence of adversarial examples~\cite{goodfellow2014explaining} highlights the vulnerability of FNNs and raises concerns about their safety in critical domains. Adversaries can easily deceive FNNs by introducing small and imperceptible perturbations to natural inputs, resulting in erroneous predictions. Numerous studies have attempted to enhance and verify the robustness of FNNs. 
Although Adversarial training~\cite{madry2017towards} is currently considered the most effective approach for training models that are resistant to adversarial attacks, 
a serious weakness with such approaches is the lack of formal guarantees of the robustness.

To solve this problem, another avenue of research involves formally modeling and verifying the robustness of given models~\cite{huang2017safety,katz2017reluplex}.
These methods can provide provable verification of local robustness, which pertains to specific input samples. 
However, 
simply evaluating a model's local robustness against 
a test set
cannot provide global robustness analysis. To explore global robustness verification, 
\cite{ruan2019global} 
proposed to approximate the globally robust radius utilizing the Hamming distance. 
Despite this, the proposed method in \cite{ruan2019global} still depends on a test set, which is not entirely satisfactory for global robustness verification.

In addition, an inherent challenge in 
neural network verification is the computational complexity. The number of activation patterns, 
that is the potential activation status of non-linear neurons, 
can be of an exponential order of magnitude. 
Therefore, it 
is not  
practical to cover all possible patterns as the model 
size increases rapidly nowadays. To address this issue, existing approaches utilize linear relaxation~\cite{zhang2019towards} and abstract interpretation \cite{8418593}
techniques 
for adversarial training and verification.
However, these methods all focus on local robustness against the vicinity of a data point.

To achieve global robustness analysis, 
DeepGlobal~\cite{sun2021deepglobal,sun2022deepglobal} was proposed to facilitate global robustness verification of FNNs.
It
introduces a novel neural network architecture, Sliding Door Network (SDN), where 
all adversarial regions can be more efficiently generated.
However, the specification and implementation of 
the proposed neural network SDN in DeepGlobal~\cite{sun2021deepglobal,sun2022deepglobal} were not formally established. 
As 
rigorous formalization
is crucial for 
safety verification, 
further steps must be taken to ensure that 
the global robustness 
of the new 
neural network 
SDN
can be
formally 
proven.

In this paper, we build upon the DeepGlobal framework and use the SMT solver Z3~\cite{moura2008z3} to provide a complete specification and implementation of the framework. Specifically, we provide 
formal definitions 
of DeepGlobal and algorithms with several improvements for more efficient generation of adversarial dangerous regions. 
To 
demonstrate 
how the Z3 solver can be applied to verify the global robustness of FNNs,
we conduct extensive experiments on the MNIST~\cite{lecun1998mnist} and FashionMNIST~\cite{xiao2017fashion} datasets. The code is available at \url{https://github.com/weizeming/Z3_for_Verification_of_FNN_Global_Robustness}.

In summary, our contributions in this paper are:
\begin{enumerate}
    \item We provide a complete specification and implementation of the DeepGlobal framework by utilizing the Z3 SMT solver.
    \item We propose several improvements to the original DeepGlobal framework, including more explicit definitions and more efficient verification algorithm.
    \item We conduct experiments on a set of benchmark datasets to demonstrate the validity and effectiveness of our  implementation.
\end{enumerate}

The paper is organized as follows. In Section~\ref{pre}, we provide preliminaries on Feedforward Neural Networks (FNNs), Adversarial Dangerous Regions (ADRs) and Sliding Door Activation (SDA). We introduce the Z3 specification for FNNs and Sliding Door Networks (SDNs) proposed in DeepGlobal in Section~\ref{SDN}, which is used in this robustness verification framework. 
In Section~\ref{rule}, we further show
the Z3 specifications of SDNs and ADRs, which provides an explicit definition of the DeepGlobal framework. Furthermore, we present algorithmic implementation details in Section~\ref{imple}. We report the experiment results on benchmark datasets in Appendix~\ref{exp}.

\section{Preliminaries}
\label{pre}
\subsection{Feedforward Neural Networks}
We consider a $K$-classification 
neural network
$F:\mathcal X\to\mathcal Y$, which maps an input space $\mathcal X\subset \mathbb R^d$ to an output space $\mathcal Y=\{1,2,\cdots, K\}$. Let $\tilde F(x)$ denote the ground-truth classification result for $x\in\mathcal X$ as determined by a human expert.

We define a feedforward neural network (FNN) $f$ as a tuple $(m, N, W, A)$, where $m$ is the number of layers in $f$, $N=(n_1, n_2, \cdots, n_m)$ is a vector specifying the number of neurons in each layer, $W=(w_1, b_1,\cdots,w_m,b_m)$ is a set of parameters for $f$, where $w_i\in\mathbb R^{n_i\times n_{i-1}}$ and $b_i\in\mathbb R^{n_i}$, and $A=(a_1,a_2,\cdots, a_m)$ is a set of activation functions for each layer, where $a_i:\mathbb R^{n_i}\to \mathbb R^{n_i}$. Thus, the function computed by $f$ can be expressed as
\begin{equation}
f(x) = a_{m}(w_m\cdots  (w_1\cdot x + b_1))+b_m).
\end{equation}

Note that the input dimension of the FNN $f$ satisfying that $n_0 = d$, and the output dimension  $n_m = K$. Given $f(x) = (f(x)_1,f(x)_2,\cdots, f(x)_K)$, the FNN returns its prediction $F(x) = \arg\max\limits_i f(x)_i$.


\subsection{Adversarial Examples and Dangerous Regions}
\textit{Adversarial examples}~\cite{szegedy2013intriguing,goodfellow2014explaining} are inputs that have a small perturbation $\delta$ added to a benign sample $x$ such that the model misclassifies the perturbed sample $F(x+\delta) \neq F(x)$.
Typically, the perturbation $\delta$ is constrained by an $l_p$-norm ball, such as $|\delta|_p \le\epsilon$. 

The concept of \textit{Adversarial Dangerous Regions} (ADRs)
is 
introduced to characterize global robustness. 
ADRs characterize 
the potential regions where the model's prediction is near the decision boundary and the samples in it has clear semantics. We can formally model these conditions as
\begin{equation}
\begin{split}
    \label{eq:adr}
    &Ad_{F} := \{x|\ \exists y, i\ne j: \|x-y\|_p \le \epsilon,\\ F(y)_i = &F(y)_j \ge F(y)_k (\forall k\ne i,j), \tilde F(y) = i\}.
\end{split}
\end{equation} 


\subsection{Sliding Door Activation (SDA)}
\label{sda}
In the DeepGlobal framework, the sliding door activation (SDA) function is proposed to reduce the 
number 
of activation patterns in the new type of 
neural networks
(SDNs). SDA divides the neurons in each layer $(\hat h_{i,1}, \hat h_{i,2}, \cdots, \hat h_{i,n_i})$ 
into groups of $k$ neurons. Let the divided groups be denoted as $G_{h, 0}, G_{h,1}, \cdots, G_{h, l}$, where $l = \frac{n_i}{k}$. SDA finds the first group $G_{h, Act}$, in which pre-activated neurons are all positive, from left to right. This group presents the property that each neuron within it is active and is preferred for activation. Therefore, SDA names this group the \textit{Active Door} and multiplies it by a constant $\alpha > 1$ to stimulate the active neurons as activation. Additionally, SDA searches for an \textit{Inactive Door} $G_{h, Ina}$ in which neurons are all negative and multiplies them by 0 to penalize the inactive neurons. The remaining $l-2$ doors are named \textit{Trivial Doors}, which SDA neither activates nor deactivates but retains their values after activation. SDN leverages SDA to achieve comparable accuracy to general FNNs, such as networks with ReLU activation function, while significantly reducing the magnitude of activation patterns, making it an efficient candidate for verification.

\section{Formalization of Sliding Door Networks}
\label{SDN}

\subsection{Formulation of FNNs}
\label{fm}
As the concept of SDNs is based on FNNs, we 
first demonstrate how to use the SMT solver Z3~\cite{moura2008z3} to formally model a given FNN in this section.
We assume
the FNN configuration (e.g., input dimension $d$) has already been declared.

To start, we represent each variable in the input, hidden, and output layers as a `Real' object in Z3:

\begin{lstlisting}[%title={Declaration of Variables},
basicstyle=\small]
Input = [Real(f"x_{i}") for i in range(d)]
Hidden = [[Real(f"h_{i}_{j}") for j in range(N[i])] for i in range(m-1)]
Output = [Real(f"y_{i}") for i in range(K)]
\end{lstlisting}

In this way, the input and output variables are named ‘$x\_i$', ‘$y\_i$' respectively, where $i$ indicates the $i$-th input (output) variable (counting from zero). For $0\le i\le m-2$, the $j$-th hidden variable in the $i$-th layer is named `$h\_i\_j$' (counting from zero), and note that the $m-1$-th layer is the output layer. 

Next, the constraints between input, hidden, and output layers can be modeled. Note that the constraint relations are highly dependent on the activation patterns. Therefore, we can only model the constraints for each potential activation pattern respectively. 
The constraints include four parts:
\begin{enumerate}
    \item \textbf{The constraints on input domain $\mathcal X$.} Taking the MNIST dataset as example, since each pixel value is restricted to $[0,1]$, we have
    \begin{lstlisting}[%title={Constraints on input domain},
    basicstyle=\small]
s = Solver()
s.add([Input[i] >= 0 for i in range(d)])
s.add([Input[i] <= 1 for i in range(d)])
    \end{lstlisting}
    where $s$ is the initialized solver to be used later. We denote these constraints as $C_{Input}$.
    
    \item \textbf{The relation between adjacent layers under given activation patterns.} The forward-pass from $h_{i-1}$ (the $i-1$-th layer) to $h_i$ (the $i$-th layer) can be formulated as $h_i = a_i(w_i\cdot h_{i-1}+b_i)$. 
    For the sake of simplicity, we 
    introduce variables `$\_h\_i\_j$' for
    the pre-activate neurons $\hat h_i = w_i\cdot h_{i-1}+b_i$:
\begin{lstlisting}[% title={Declaration of Pre-activate neurons}, 
basicstyle=\small]
_Hidden = [[Real(f"_h_{i}_{j}") for j in range(N[i])]
            for i in range(m-1)]
\end{lstlisting}
In this way, we can simplify the constraint from layer $h_{i-1}$ to $h_i$ with the aid of $\hat h_i$:
\begin{lstlisting}[%title={Constraints of Pre-activate neurons}, 
basicstyle=\small]
s.add(_Hidden[i][j] == Sum([W[i][j][k] * Hidden[i-1][k] for k in range(n_{i-1})]) + B[i][j])
s.add(Hidden[i][j] == a[i](_Hidden[i][j])) //pseudo-code
\end{lstlisting}
    Here we use pseudo-code to show the constraint of activation function $a[i]$.
    The details of activation functions are introduced in Section~\ref{sda} 
    and its Z3 specification is presented in Section~\ref{rule}. 
    
    \item \textbf{The activation condition of  the given activation patterns.} We defer this part in Section~\ref{rule}  after we introduce the Sliding Door Activation (SDA) functions.
    
    \item \textbf{The objective property.} For example, if we want to identify samples from class $i$, 
    which are also near the decision boundary with class $j$, the constraints should be formulated as $f(x)_i = f(x)_j$ $\bigwedge_{k\ne i,j} f(x)_i\ge f(x)_k$. This can be expressed with Z3 constraints as:
    
\begin{lstlisting}[%title={Constraints of decision boundary},
basicstyle=\small]
s.add(Output[i] == Output[j])
for k in range(K):
    if k == i or k == j:
        continue
    s.add(Output[i] >= Output[k])
\end{lstlisting}
The  specification details of adversarial dangerous regions (ADRs) is presented in Section~\ref{rule}. 
\end{enumerate}

\subsection{Formulation of SDNs}

We now provide the complete definition of Sliding Door Networks (SDNs). A SDN is a feedforward neural network $f$ with the tuple $(m,N,W,A,k)$, where $m$, $N$, and $W$ are inherited from the definition of a FNN, and $A=(a_1,\cdots,a_m)$ 
 is the sliding door activation function. The parameter $k$ represents the number of neurons in each group. Let $h_i$ denote the hidden variables in the $i$-th layer, with 
$h_0=x$ 
being the input and $\hat h_{m+1}=f(x)$ being the output. We can recursively define the mapping $f$ as follows:

\begin{equation}
\label{eq:sdn}
\begin{split}
    &\hat h_i = w_i\cdot h_{i-1} + b_i,\\
    &\begin{cases}
    Act_i = \arg\min\limits_{g}\ \forall (g-1)\cdot k < j\le  g\cdot k, &\hat   h_{i,j}>0,\\
    Ina_i = \arg\min\limits_{g}\ \forall (g-1)\cdot k < j\le  g\cdot k, &\hat   h_{i,j}<0.\\
    \end{cases}\\
    &
    h_{i,j} = 
    \begin{cases}
    \alpha\cdot\hat  h_{i,j} , &(Act_i-1)\cdot k < j\le  Act_i\cdot k\\
    0, &(Ina_i-1)\cdot k < j\le  Ina_i\cdot k\\
    \hat h_{i,j}, & else,
    \end{cases}
\end{split}
\end{equation}

Note that the $Act_i$ or $Ina_i$ in \eqref{eq:sdn} may not exist in some layers. In this case, SDN simply abandons the active or inactive door when mapping through these layers.
We further discuss the computational cost of enumerating all 
Activation Patterns in Appendix~\ref{apa}, which demonstrate superiority the than verifing on classic FNNs.

\section{Complete Modeling}
\label{rule}

\subsection{Modeling the activation conditions}
\label{ac}
As discussed in Section~\ref{fm}, the specification of FNNs depends on the Activation Patterns (AP), i.e., the different configurations of active and inactive neurons in the network. For a SDN with $m$ layers, we define an activation pattern $\mathcal A=(Act_1, Ina_1,\cdots, Act_m, Ina_m)$, where $Act_i$ and $Ina_i$ correspond to the indices of the active and inactive doors in layer $i$, respectively (counting from 0 to be consistent with Python code). If the active or inactive door does not exist, we fill $Act_i$ or $Ina_i$ with $\frac{n_i}{k}$, which is the number of groups in this layer.

Therefore, given an activation pattern $\mathcal A$, we give the specification of activation conditions as:
\begin{lstlisting}[%title={Constraints of Activation Conditions},
basicstyle=\small]
for i in range(m):
    if Act[i] != n[i]//k:
        s.add([_Hidden[i][j] > 0 for j in range(Act[i] * k, (Act[i]+1) * k)])
    if Inc[i] != n[i]//k:
        s.add([_Hidden[i][j] < 0 for j in range(Ina[i] * k, (Ina[i]+1) * k)])
\end{lstlisting}

We denote the constraints described above as $C_{AP}(\mathcal A)$. When $Act[i]$ or $Ina[i]$ is equal to $\frac{n_i}{k}$, we skip this constraint. Note that we do not explicitly model the minimality of $Act[i]$ or $Ina[i]$, which may result in covered and common boundaries of activation regions.

The above
issues are addressed by successively eliminating already-covered or common boundaries in~\cite{sun2021deepglobal,sun2022deepglobal}. 
For instance, to remove covered or common boundaries with a previous region $\bigwedge P_j$, they conjunct each $\lnot P_j$ with $\bigwedge_i R_i$ to create a new region. 
Using this approach, 
we only need to consider $\lnot C_{AP}(\mathcal A')\land C_{AP}(\mathcal A)$ to remove covered and common boundaries with $\mathcal A'$ for $\mathcal A$. The complete algorithm is presented in Algorithm~\ref{alg}.

\subsection{Modeling Sliding Door Activation}
Recall from Section~\ref{fm} that we have modeled the linear transformation from $h_{i-1}$ to $\hat{h}_i$. Now, we provide the formal specification of the activation function $h_i = a_i(\hat{h}_{i-1})$, which is dependent on a given activation pattern $\mathcal{A}$.

\begin{lstlisting}[%title={Specification of Sliding Door Activation},
basicstyle=\small]
for i in range(m):
    for j in range(n[i]//k):
        if Act[i] == j:  # Active Door
            s.add([Hidden[i][j+l] == alpha * _Hidden[i][j] for l in range(k)])
        elif Ina[i] == j:  # Inactive Door
            s.add([Hidden[i][j+l] == 0 for l in range(k)])
        else:  # Trivial Door
            s.add([Hidden[i][j+l] ==  _Hidden[i][j] for l in range(k)])
\end{lstlisting}

We denote this set of constraints (including the constraints on linear mappings) as $C_{Forward}(\mathcal A)$.

\subsection{Modeling the Adversarial Dangerous Regions}
\label{modeling adr}
Recall our refined definition of Adversarial Dangerous Regions in Section~\ref{pre}, where we aim to find feasible $y$ that satisfies the \textit{boundary condition} (i.e., $\exists i \ne j$ such that $\forall k \ne i,j.\, F(y)_i = F(y)_j \ge F(y)_k$) and the \textit{meaningful condition} (i.e., $\tilde F(y) = i$). 
In Section~\ref{fm}, we present the Z3 specification for the boundary condition, which we denote as $C_{Boundary}(i,j)$. 
\cite{sun2021deepglobal,sun2022deepglobal} do not consider the meaningful condition. Instead, they attempt to find feasible and meaningful solutions in the ADRs. 
Specifically, they use a trained autoencoder~\cite{bank2020autoencoders} to optimize a feasible solution $x^0$ in a given ADR, while ensuring that it remains within the same ADR. However, this optimization-based method has several limitations. For instance, the meaningful solution may not always exist for all ADRs,
which is a possible scenario 
when all samples in the region are deemed "rubbish". Additionally, optimizing the solution along certain directions within the region can be extremely time-consuming.

Therefore, we propose a new approach that allows for more straightforward identification of meaningful samples. Note that the \textit{meaningful condition} is $\tilde F(y) = i$. While judging each sample by $\tilde F$ (\textit{i.e.}, human-perception) is not practical, we can still use autoencoders as surrogate models.

For a given class $i\in\{1,2,\cdots, K\}$, we hope to find a meaningful region by the surrogate model $AE$ where $\tilde F(y)=i$. 
To achieve this, we train an autoencoder $E(\cdot)$ and leverage it to define the center of class $i$ as $c_i = \frac {1}{|X_i|} \sum_{x\in X_i} E(x)$, where $E(\cdot)$ is the encoder function, and $X_i$ represents the samples in the training set with class $i$, and define the \textit{prototype} of class $i$ as $P_i = D(c_i)$. The prototype model for class $i$ is decoded from the average code of samples in that class, making it a standard representation of that class. Our assumption is that any meaningful sample $y$ with $\tilde F(y) = i$ should not be significantly different from the prototype $P_i$. To ensure this, we restrict $y$ to a \textit{meaningful region} $|y - P_i|_p\le r$, where $r$ is a pre-specified radius. It's worth noting that the definition of the meaningful region is fundamentally different from that of adversarial examples (see Section~\ref{pre}), where the perturbation $\delta$ is limited to a specific bound $\epsilon$.
Generally, $r$ is much larger than $\epsilon$, as all samples in this region are close to the prototype and potentially meaningful. The definition of adversarial examples is more restrictive than that of meaningful regions, as it only focuses on a small perturbed region.

Based on the above analysis, taking $l_\infty$-norm as example, we 
specify the meaningful condition as follows:
\begin{lstlisting}[%title={Specification of Meaningful Condition},
basicstyle=\small]
for i in range(d):
    s.add([Input[i] - P[i] < r, P[i] - Input[i] < r])
\end{lstlisting}
We denote this set
of constraints as $C_{Meaningful}(i)$. 
So far, we have completed all specifications for the DeepGlobal verification framework in Z3. 
To find the feasible and meaningful sample $y$ in class $i$ (referred to as \textit{target class}), which is on the decision boundary of class $j$ (referred to as \textit{boundary class}), with regard to activation pattern $\mathcal A$, one only need to solve the following constraints in Z3:
\begin{equation*}
\begin{split}
        &C_{Input}\land C_{AP}(\mathcal A)\land  C_{Forward}(\mathcal A)\land\\& C_{Boundary}(i,j)\land C_{Meaningful}(i).
\end{split}
\end{equation*}

\section{Algorithmic Implementation Details}
\label{imple}
In this section, we demonstrate the implementation of using Z3 solver to specify the DeepGlobal framework and identify global adversarial regions.
\subsection{Finding solutions for target and boundary classes}
To find samples $y$ that belong to class $i$ and are on the decision boundary of class $j$ (\textit{i.e.}, $F(y)_i=F(y)_j\ge F(y)_k(\forall k\ne i,j)$), we need to enumerate all target-boundary class pairs $(i,j)$. These samples are referred to as \textit{boundary samples}, which form a complete set of samples that support the adversarial dangerous regions.

Algorithm~\ref{alg} presents a complete workflow for this implementation. Line 1 initializes the input, boundary, and meaningful constraints, which are shared for each valid activation pattern. In line 2, we use $C_{checked}$ to track the regions that have already been checked in previous activation patterns to avoid redundancy, as described in Section~\ref{ac}. The $Solutions$ list in line 3 stores the solved feasible and meaningful samples.
In lines 4-8, we create a Z3 solver $s$ for each valid activation pattern $\mathcal A$ and add the required constraints to it. If the constraints can be solved, we append the 
generated
sample to $Solutions$ as shown in lines 9-11. Then, we add the checked region to $C_{checked}$ in line 12 to avoid solving it again for
other activation patterns.
Finally, the algorithm returns all feasible and meaningful solutions for the target class $i$ and boundary class $j$. 
\begin{algorithm}[t]
\label{alg}
  \SetAlgoLined
  \KwIn{SDN Network $f=(m, N, W, A, k)$; Target class $i$; Boundary class $j$}
  \KwOut{Feasible and Meaningful solutions}
    $Initialize\quad  C_{Input},\ C_{Boundary}(i,j),\ C_{Meaningful}(i)$\;
    $C_{Checked}\gets\emptyset$\;
    $Solutions\gets \emptyset$\;
    \For{All valid AP $\mathcal A$}{
        $s\gets new\ Z3\ solver$\;
        $s.add([C_{Input},\ C_{Boundary}(i,j),\ C_{Meaningful}(i)])$\;
        $s.add(\ \lnot C_{Checked}\ )$\;
        $s.add([C_{AP}(\mathcal A),\ C_{Forward}(\mathcal A)]$\;
        \If{$s.solve() == sat$}{
            $Solutions.Append(s.model())$\;
        }
        $C_{Checked} \gets C_{checked}\lor C_{AP}(\mathcal A)$\;
    }
    \textbf{return} $Solutions$\;
  \caption{Find feasible and meaningful solutions}
\end{algorithm}

\subsection{Enumeration of activation patterns}
In this section, we discuss the details for implementing enumeration of activation patterns in line 4 of
Algorithm~\ref{alg}.
Recall that there are $\frac{n_i}{k}+1$ possible values for $Act_i$ and $Ina_i$, respectively. 
The only one constraint on $Act_i$ and $Ina_i$ is $(Act_i\ne Ina_i)\lor Act_i = \frac{n_i}{k}$, since any group cannot be both active and inactive door, except one case that $Act_i = Ina_i = \frac{n_i}{k},$ \textit{ i.e.}, the groups are neither activated nor inactivated. 
We 
arrange all activation patterns in a tree structure. In this way, we can implement the enumeration of activation patterns 
by the breadth-first search (BFS) algorithm
and execute from the shallow layers to the deep layers,
which is detailed in Appendix~\ref{enu}.
\subsection{Experiments}
The experiment includes two parts: the utilization of autoencoders and the generation of boundary and adversarial examples. Autoencoders are employed to generate prototypes for each dataset that represent meaningful samples with explicit semantics. The prototypes can be used for global verification, distinct from instance-wise local robustness. Further details of the autoencoders are provided in Appendix~\ref{ae}. Regarding the generation of boundary and adversarial examples, specific tactics were adapted to improve the efficiency of the Z3 solver. Boundary samples were produced for each class by identifying samples that are situated on the decision boundary between that class and the adjacent class. Adversarial examples were generated from both the exact and relaxed boundary regions. Starting from the boundary samples, perturbations were added to create adversarial examples.
Details for our experiment and created samples are shown in Appendix~\ref{exp}.

\section{Conclusion}
In this paper, we provide a complete and refined definition of the Sliding Door Networks (SDNs) and Adversarial Dangerous Regions (ADRs) in the DeepGlobal verification framework. We then present a complete specification of the framework using the SMT solver Z3 and demonstrate its detailed algorithmic implementation. Additionally, we leverage prototypes crafted by autoencoder to improve the verification framework by searching for meaningful solutions.
The experiments on two benchmark datasets show that increasing the activation coefficient $\alpha$ will lead to better model performance. 
Besides, the proposed specification support the generation of extensive boundary and adversarial samples, which can be used for identifying 
global ADRs of a given model. 
The selected customized tactics in Z3 further improve the effectiveness of our framework. 

\section*{Acknowledgement}
This research was sponsored by the National Natural Science Foundation of China under Grant No. 62172019, and CCF-Huawei Populus Grove Fund.

{
\small
\bibliographystyle{abbrvnat} 
\bibliography{ref}
\newpage
}
\appendix
\subsection{Details of Training SDNs}
\label{detail training}
\begin{figure}[!htbp]
    \centering
    \begin{tabular}{cc}
        \includegraphics[width=0.2\textwidth]{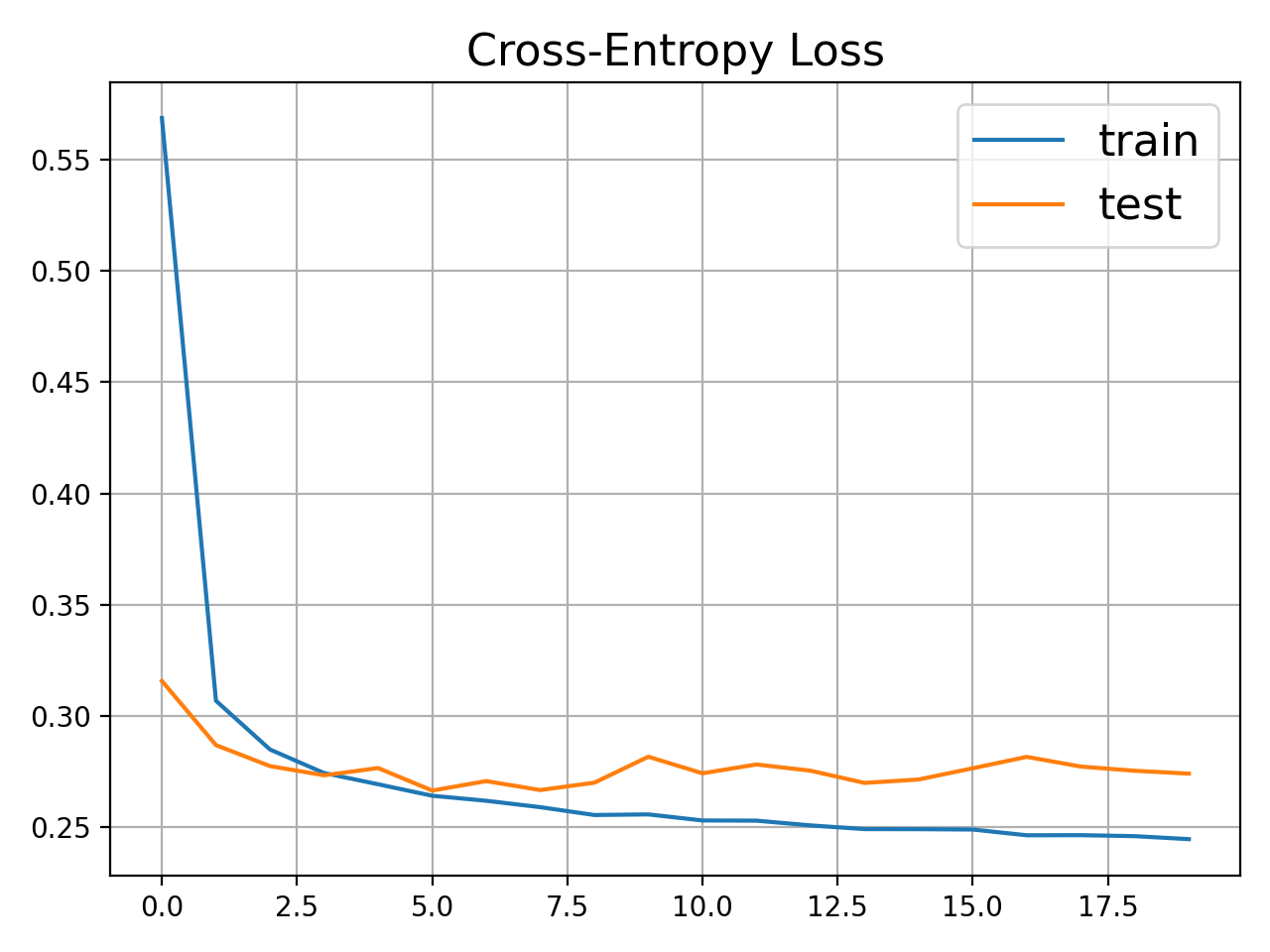}
         & \includegraphics[width=0.2\textwidth]{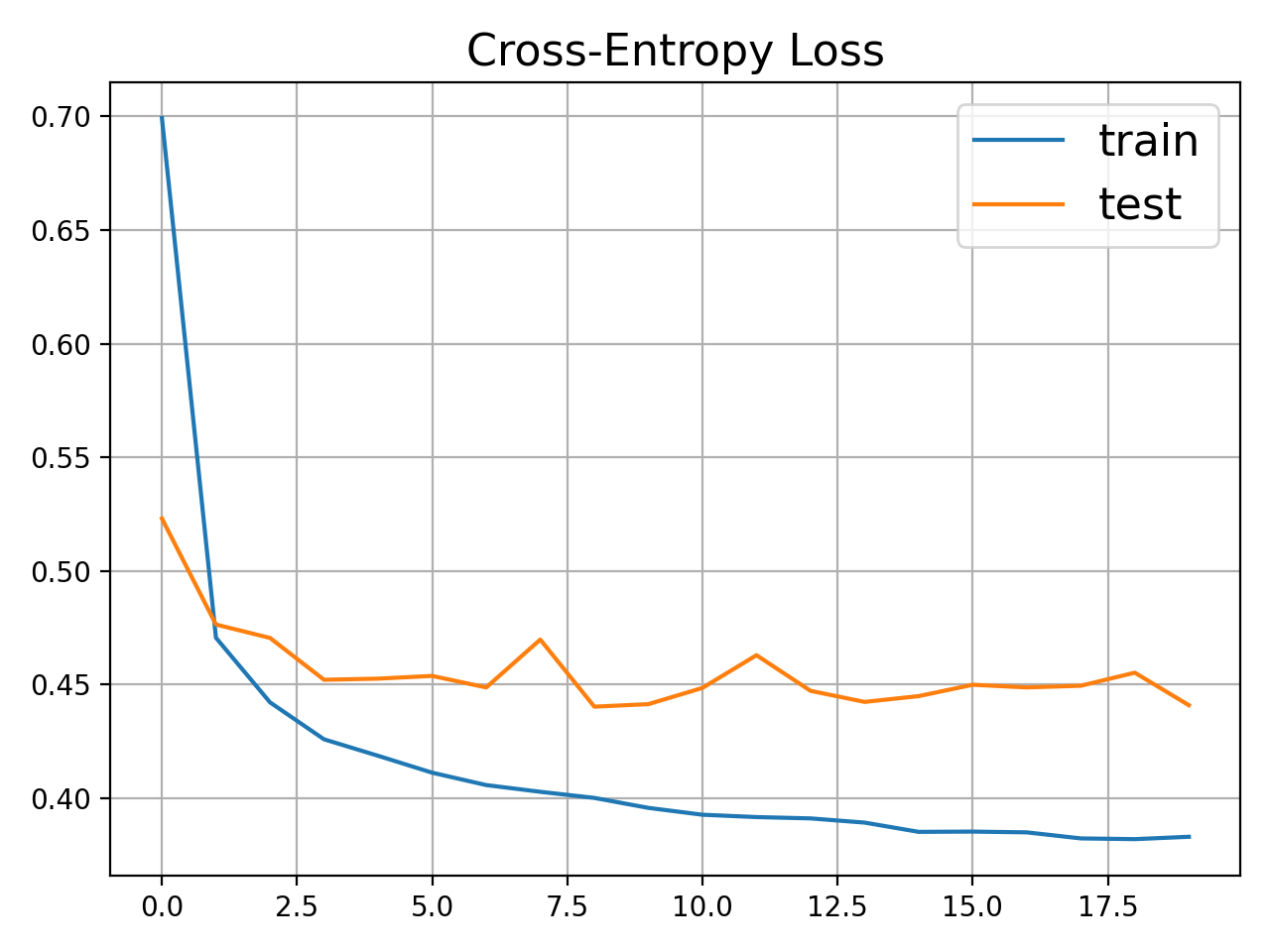} \\
        (a) Loss on MNIST  & (b) Loss on FashionMNIST \\
        & \\
        \includegraphics[width=0.2\textwidth]{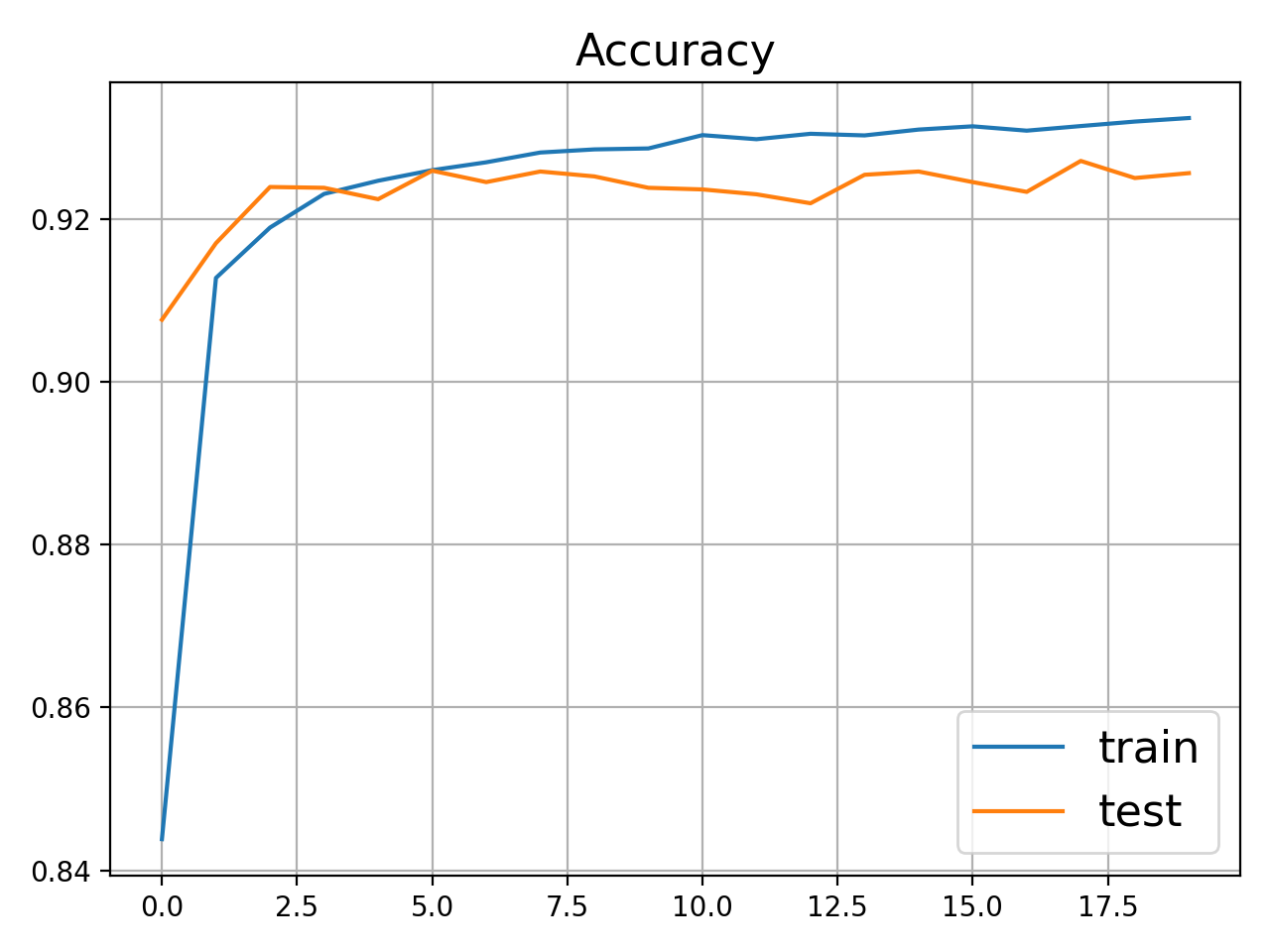}
         & \includegraphics[width=0.2\textwidth]{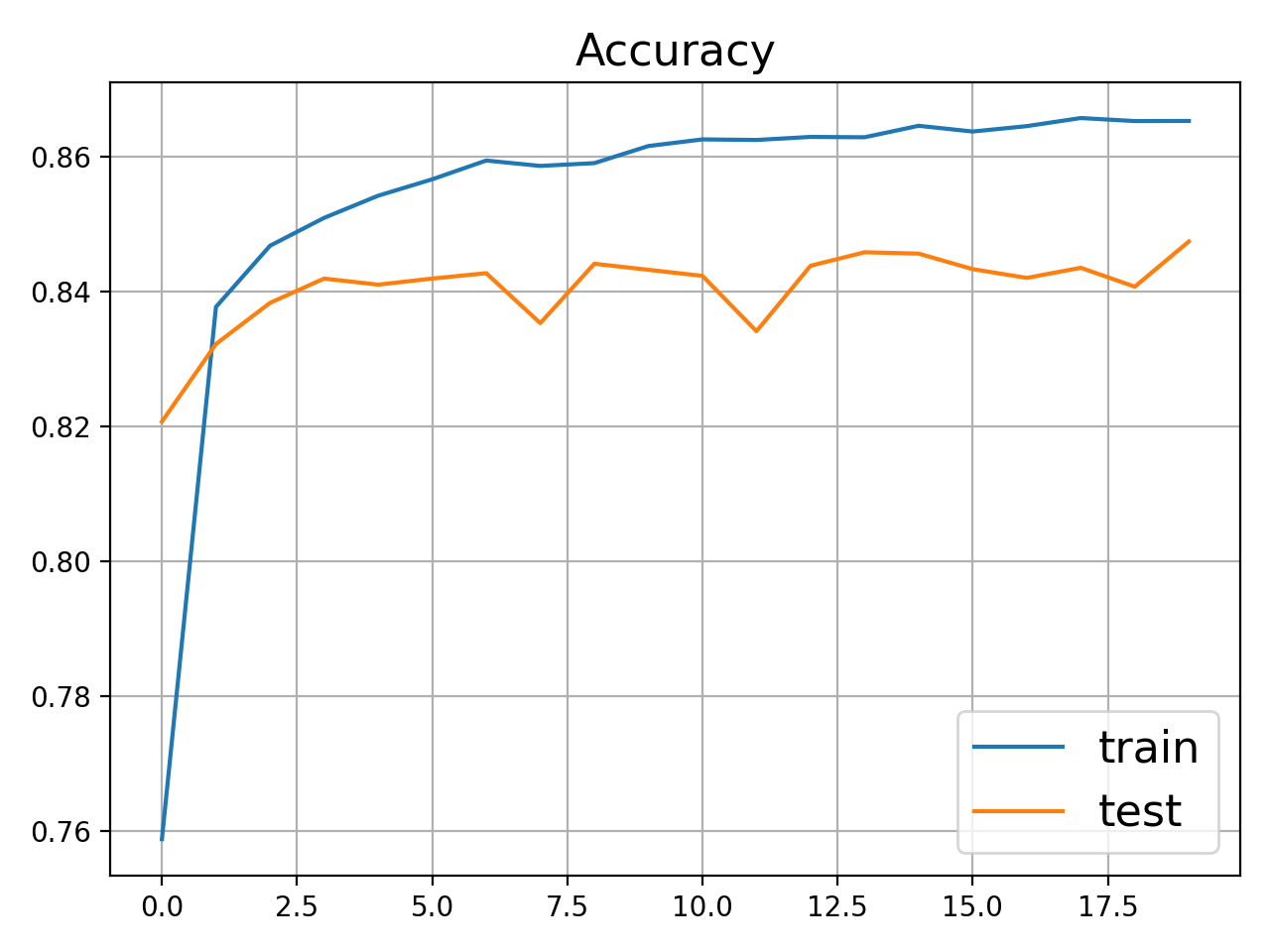} \\
       (c) Accuracy on MNIST  & (d) Accuracy on FashionMNIST \\

    \end{tabular}
    \caption{Loss and accuracy of our trained SDNs on the two datasets tracked in the training process.}
    \label{fig:train}
\end{figure}

For the both two datasets, we use the Cross-Entropy as loss function and use the Adam optimizer~\cite{kingma2014adam} to train 20 Epochs. The learning rate is selected as $0.001$ for SDN and $0.01$ for linear model, respectively. The batch size is set to be 128. The loss and accuracy for our trained SDNs is shown in Figure~\ref{fig:train}.

\subsection{Details of autoencoder}\label{ae}
An autoencoder $AE = D\circ E$ including an Encoder $E:\mathbb R^d\to\mathbb R^c$ and an decoder $D:\mathbb R^c\to \mathbb R^d$, where $c$ is the dimension of \textit{code} for input samples.
The autoencoder optimizes the following objective function:
\begin{equation}
    \mathcal L_{AE} = \mathbb E_x\ \|x - D(E(x))\|_2
\end{equation}
which makes the reconstructed sample $D(E(x))$ as similar to $x$ as possible. 
Therefore, the \textit{code} $E(x)$  can be regarded as some low-dimensional feature of $x$.

\subsection{Activation Pattern Analysis}
\label{apa}
From the perspective of activation pattern 
number SDNs demonstrate superiority than classic FNNs. For a given layer $i$ where $1\le i\le m$, SDA has no more than $(\frac{n_i}{k} + 1)^2$ possible activation patterns. This is because $Act_i$ and $Ina_i$ have $\frac{n_i}{k}$ choices each, and there is one case of non-existence. As a result, there are $\mathcal O(\prod n_i^2)$ possible activation patterns in total, which is significantly fewer than the ReLU-based FNNs that have $\mathcal O(\prod 2^{n_i})$ potential patterns. This characteristic lays a practical foundation for the global robustness verification of FNNs.

\subsection{BFS enumeration for activation patterns}
\label{enu}
See Algorithm~\ref{bfs}.
\begin{algorithm}[t]
\label{bfs}
  \SetAlgoLined
  \KwIn{SDN Network $f=(m, N, W, A, k)$}
  \KwOut{All valid activation patterns $\{\mathcal A=(Act_1, Ina_1,\cdots, Act_m,Ina_m)\}$}
    $APs\gets\{0,1,\cdots, \frac{n_1}{k}\}$\;\tcp{Initialize the first item of $\mathcal A$s}
    \While{$\mathcal A\gets APs.pop()$ \textbf{and} $\mathcal A.length < 2m$}{
    $i\gets {\mathcal A.length}//2 $\;\tcp{get the current layer of  $\mathcal A$}
    \If{$\mathcal A.length$ is odd}{
    \For{$j\ne \mathcal A[-1], 0\le j \le \frac{n_i}{k}$}{
        $APs.append(\mathcal A\cup \{j\}))$\;
    }
    \If{$\mathcal A[-1]==\frac{n_i}{k}$}{
        $APs.append(\mathcal A\cup \{\frac{n_i}{k}\}))$\;
    }

    }\Else{
    \For{$ 0\le j \le \frac{n_i}{k}$}{
        $APs.append(\mathcal A\cup \{j\}))$\;
    }
    }
    }    
    \textbf{return} $APs$\;
  \caption{BFS Enumeration for Activation Patterns}  
\end{algorithm}

\subsection{Details of experiments}
\label{exp}
\begin{figure}[!htbp]
    \centering
    \begin{tabular}{cc}
    \includegraphics[width=0.2\textwidth]{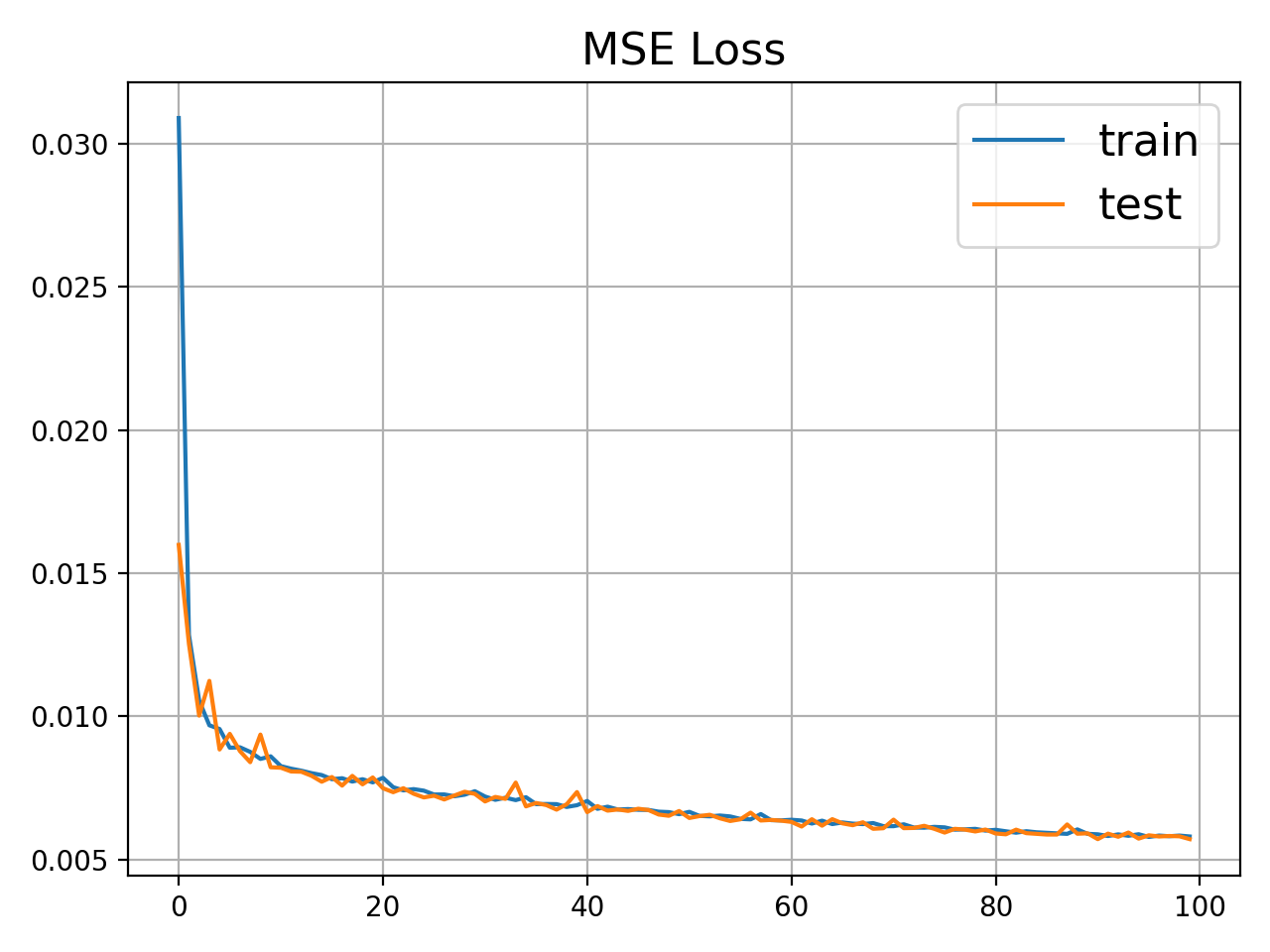}
         & 
         \includegraphics[width=0.2\textwidth]{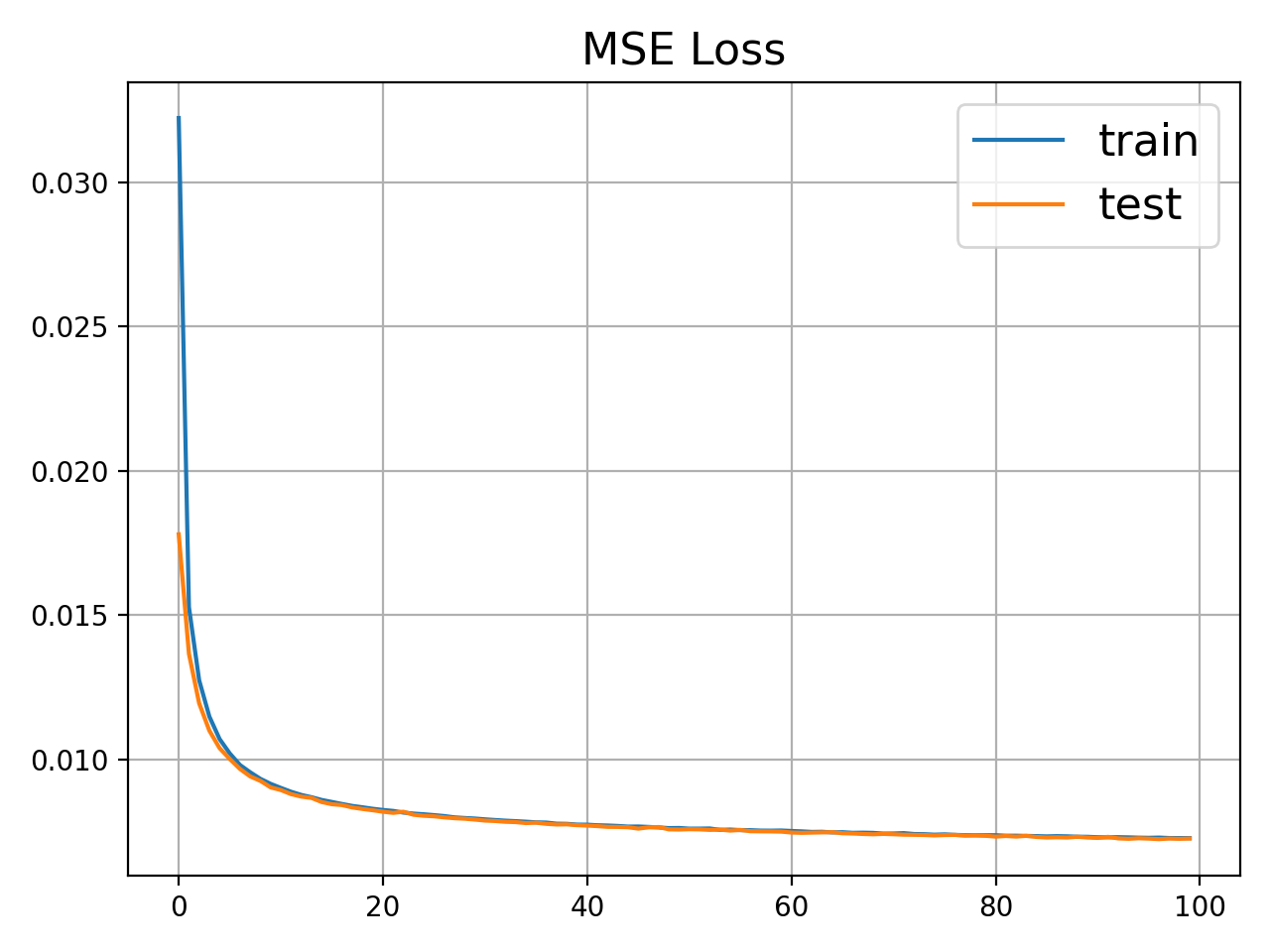}
         \\
(a) Loss on MNIST  & (b) Loss on FashionMNIST \\
\end{tabular}
    \caption{Loss  of trained autoencoders on the two datasets tracked in the training process.}
    \label{fig:ae}
\end{figure}

\textbf{Autoencoders and Prototypes.} 
\label{autoencoder}
For the two datasets, we train an autoencoder which both the encoder and decoder are two-layer ReLU FNNs, where the hidden layers have 256 neurons and the code is 64-dimension. We use mean square error (MSE) as loss function and the Adam as optimizer. The learning  rate is set to be 0.001. Both the autoencoders are trained for 100 epochs. The loss tracked in the training process is shown in Figure~\ref{fig:ae}.

\begin{figure}[!htbp]
    \centering
    \includegraphics[width=0.4\textwidth]{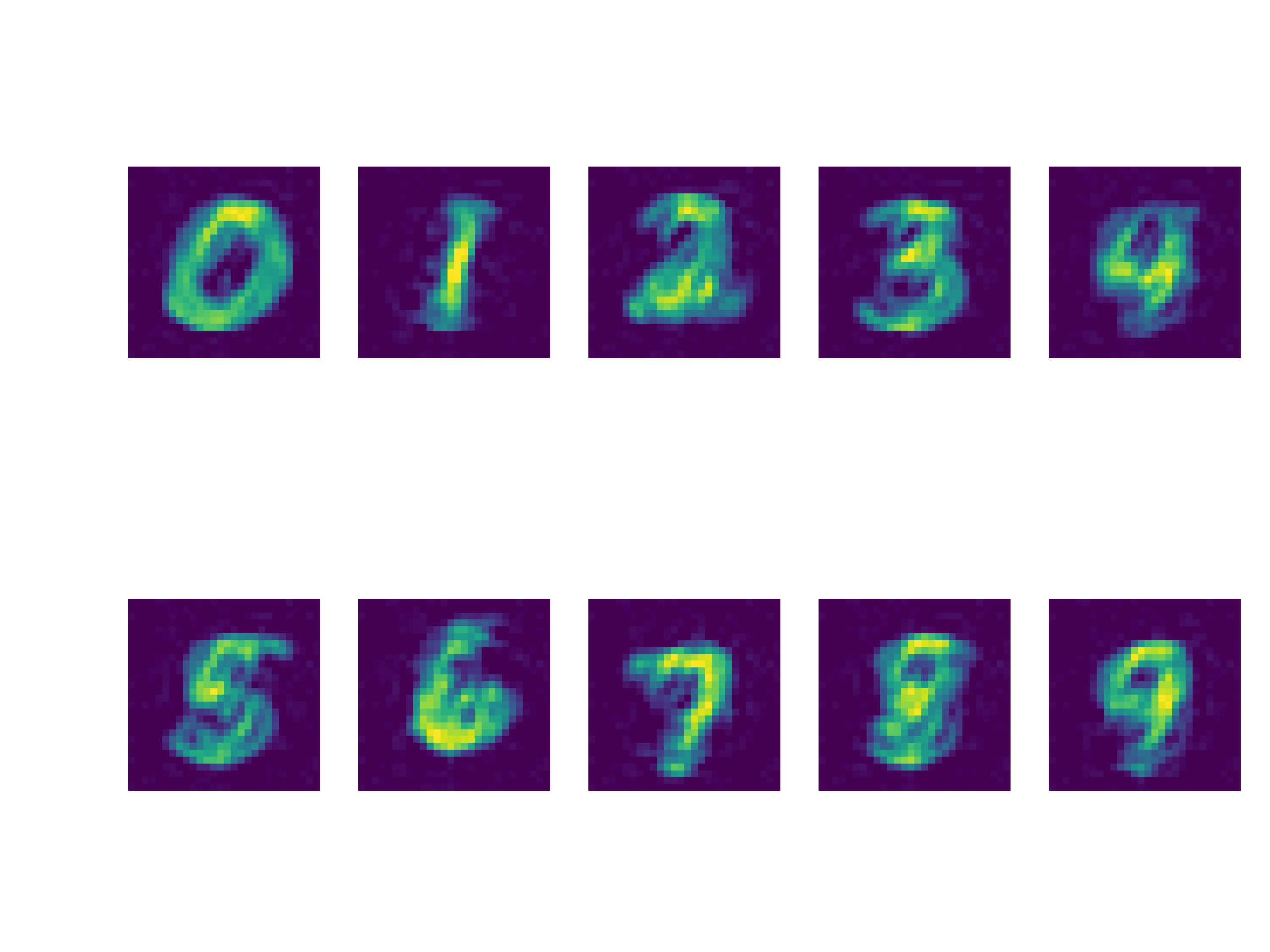}
    \caption{Visualization of all prototypes for MNIST dataset.}
    \label{fig:pro1}
\end{figure}

\begin{figure}[!htbp]
    \centering
    \includegraphics[width=0.4\textwidth]{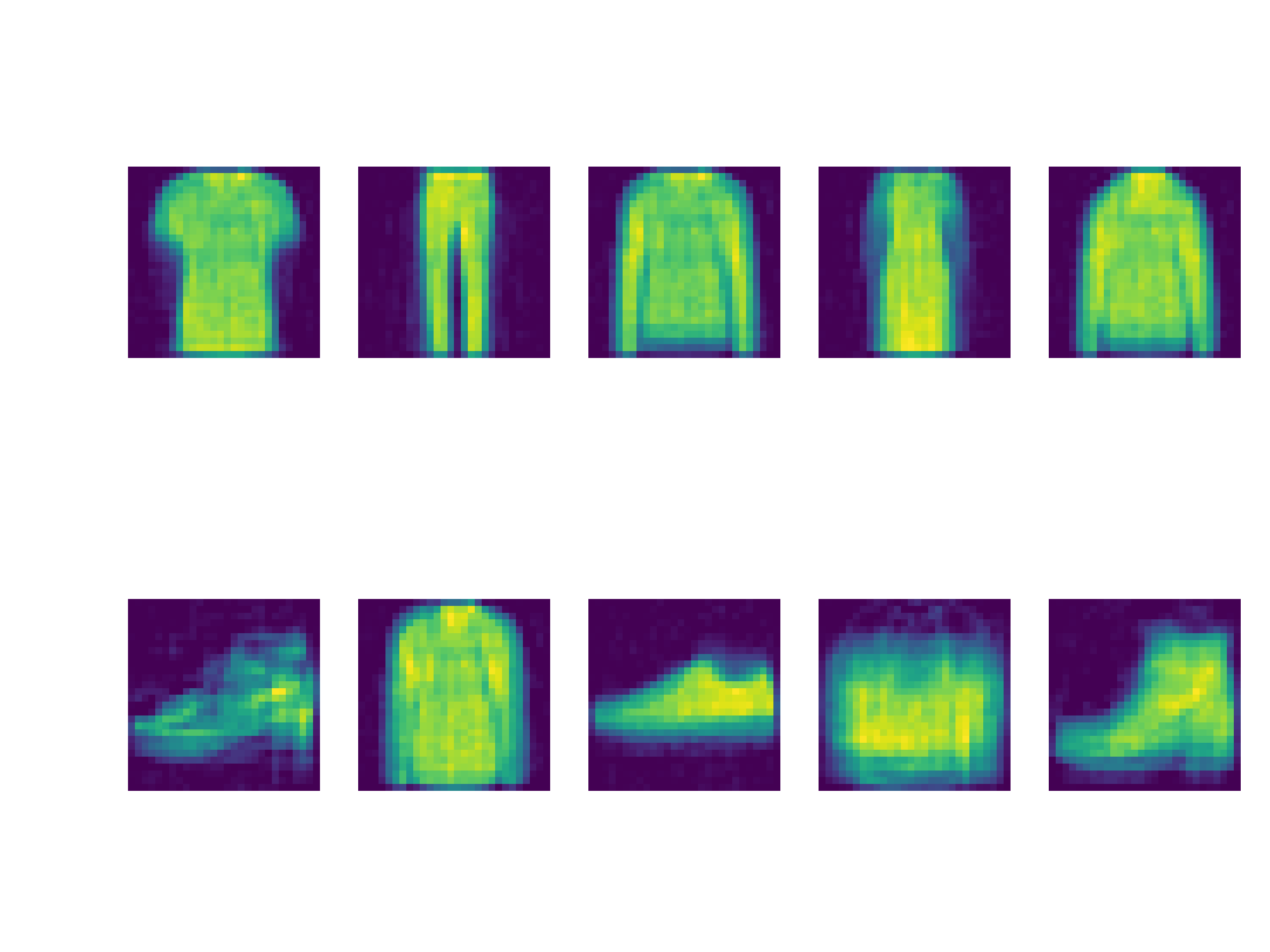}
    \caption{Visualization of all prototypes for FashionMNIST dataset.}
    \label{fig:pro2}
\end{figure}

Figure~\ref{fig:pro1} and~\ref{fig:pro2} show
some instances of the crafted prototypes for the MNIST dataset and FashionMNIST dataset respectively.

\begin{figure}
    \centering
    \includegraphics[width = 0.4\textwidth]{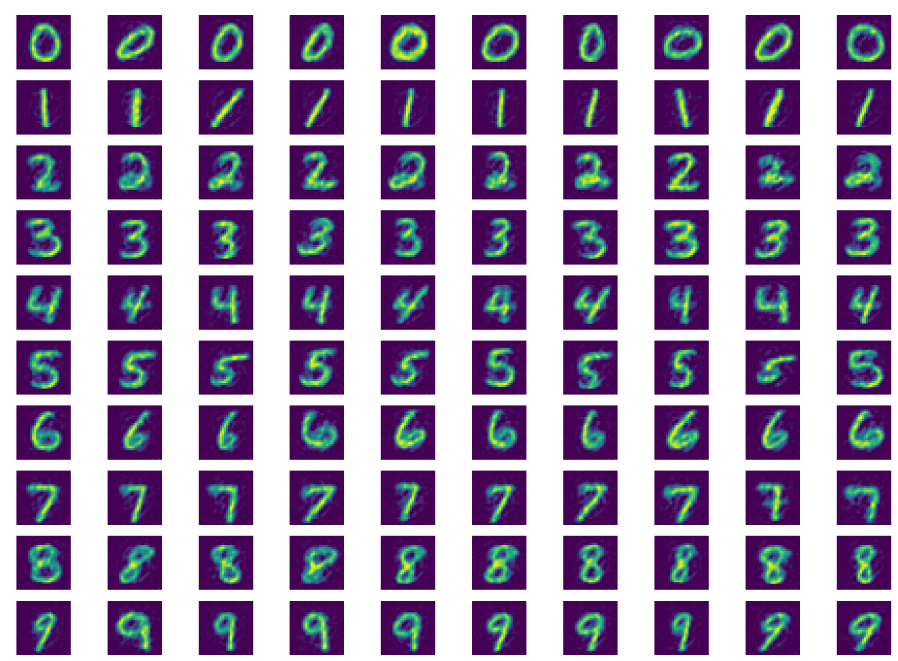}
    \caption{Visualization of all prototypes for MNIST dataset extracted by clustering. Each line shows the prototype of 10 clusters in the category.}
    \label{fig:100_pro}
\end{figure}
We observe that the prototypes in Figure~\ref{fig:pro1} and Figure~\ref{fig:pro2} represent meaningful samples and possess explicit semantics.
This highlights the fundamental difference between prototype-based global verification and instance-wise local robustness, with the former attempting to verify against all meaningful samples, while the latter focusing on specific samples.


\noindent\textbf{Boundary and Adversarial Examples.} 
To improve the efficiency of generating adversarial examples, 
we use specific tactics during initialization, including `simplify', `normalize-bounds', and `solve-eqs' .
These tactics have been customized to optimize the 
computing process for this particular task.
\begin{lstlisting}[%title={Z3 Solver Initialization with Appointed Tactics},
basicstyle=\small]
s = Then('simplify', 'normalize-bounds', 'solve-eqs','smt').solver()
\end{lstlisting}

\begin{figure}[!htbp]
    \centering
    \tabcolsep=0.01\textwidth
    \begin{tabular}{ccccc}
    \includegraphics[width=0.060\textwidth]{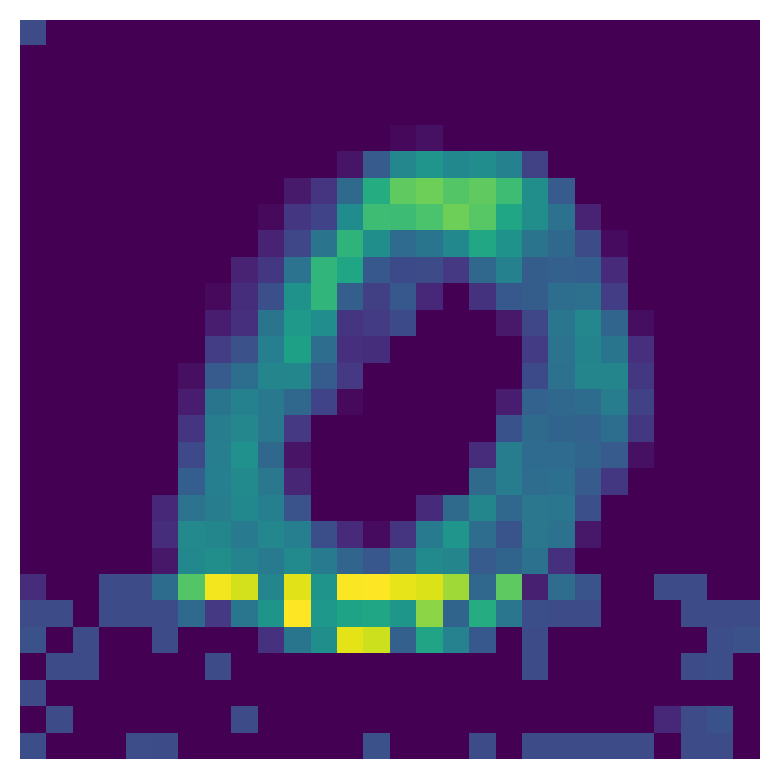} &
    \includegraphics[width=0.060\textwidth]{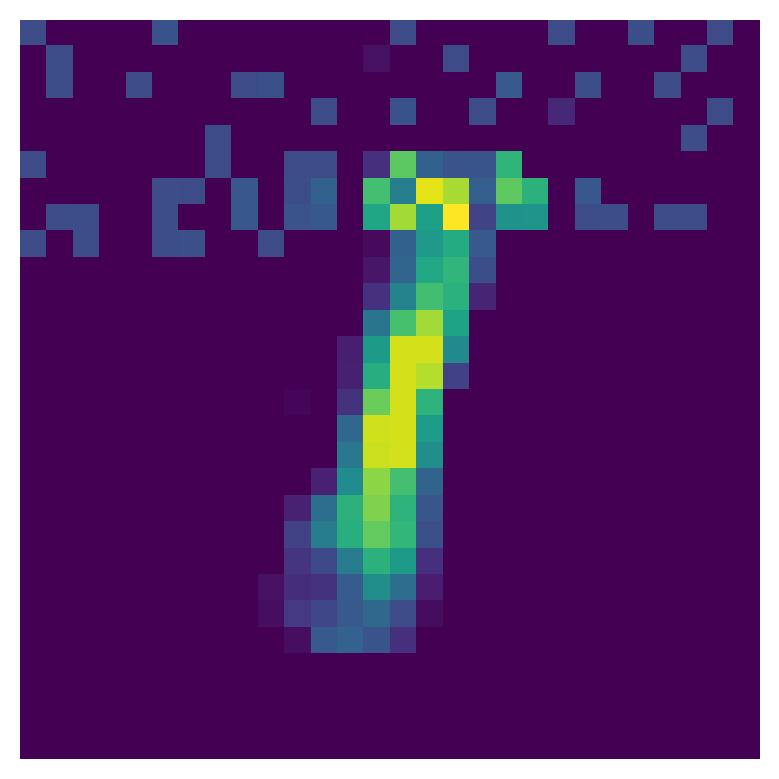} &
    \includegraphics[width=0.060\textwidth]{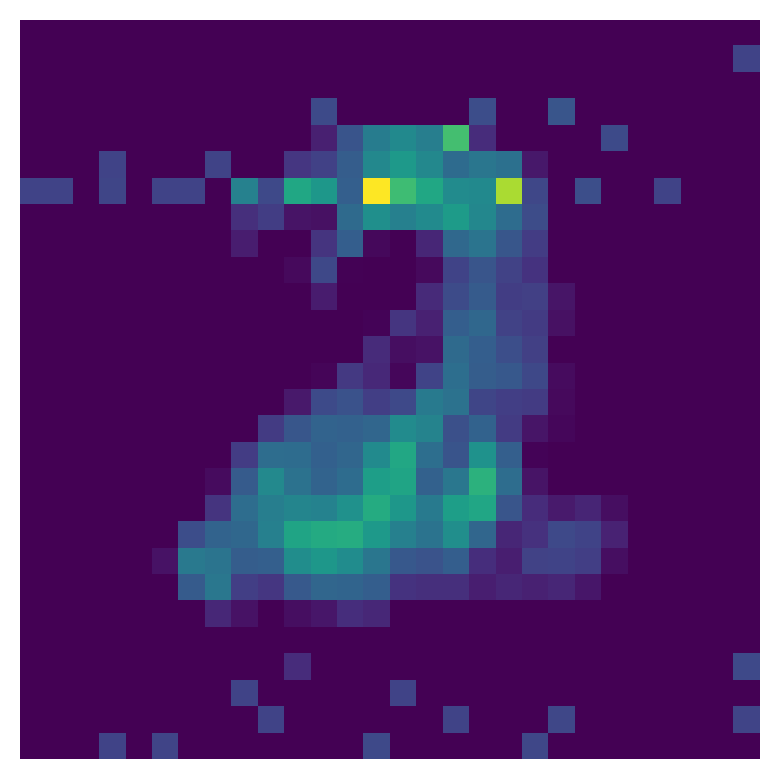} &
    \includegraphics[width=0.060\textwidth]{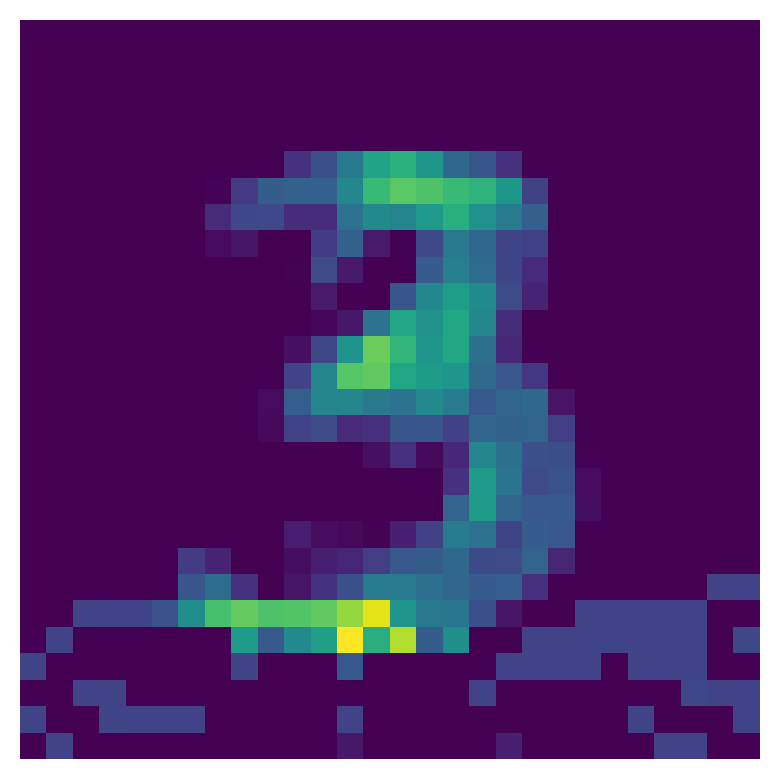} &
    \includegraphics[width=0.060\textwidth]{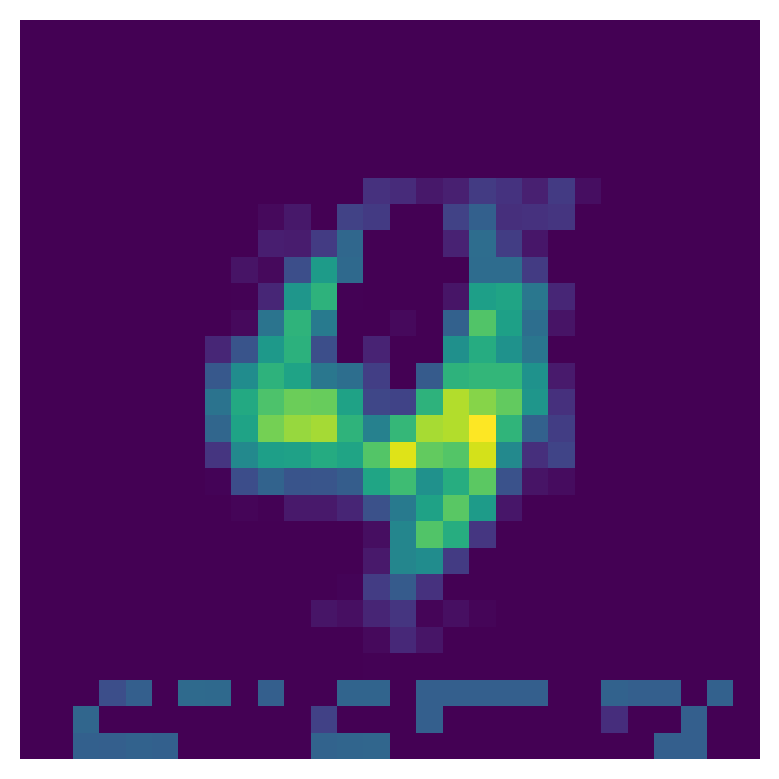} \\
    & \\
    \includegraphics[width=0.060\textwidth]{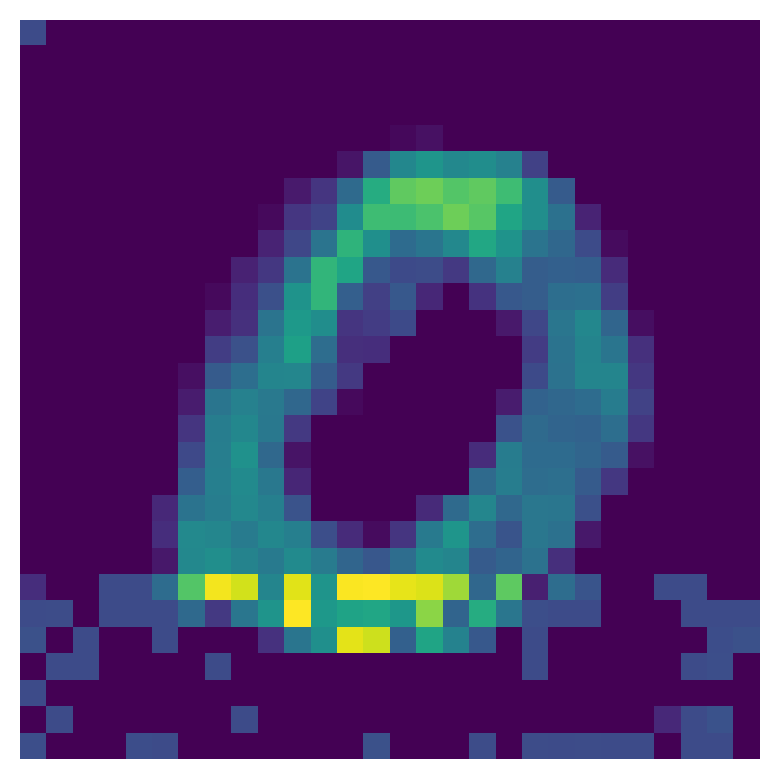} &
    \includegraphics[width=0.060\textwidth]{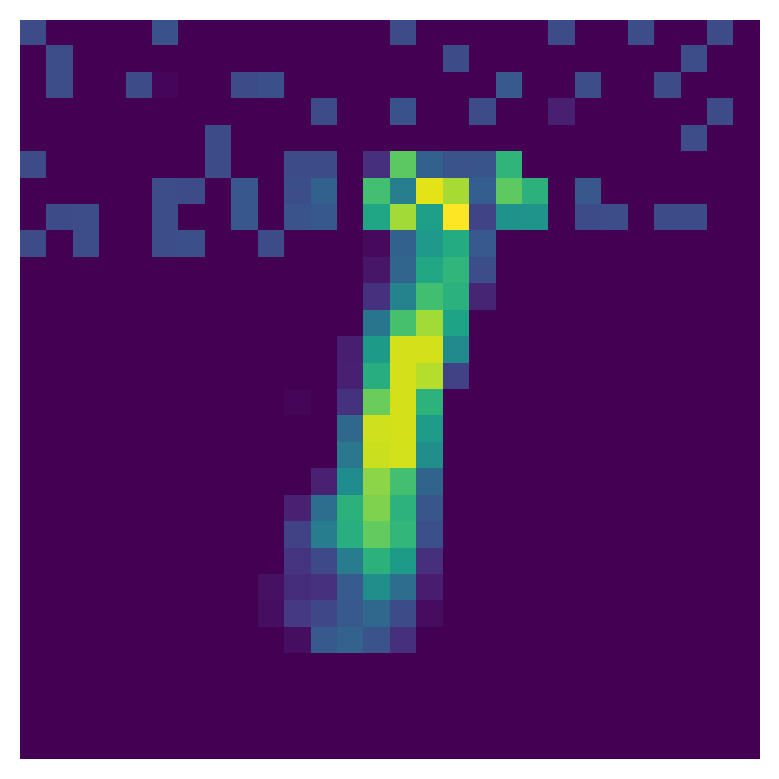} &
    \includegraphics[width=0.060\textwidth]{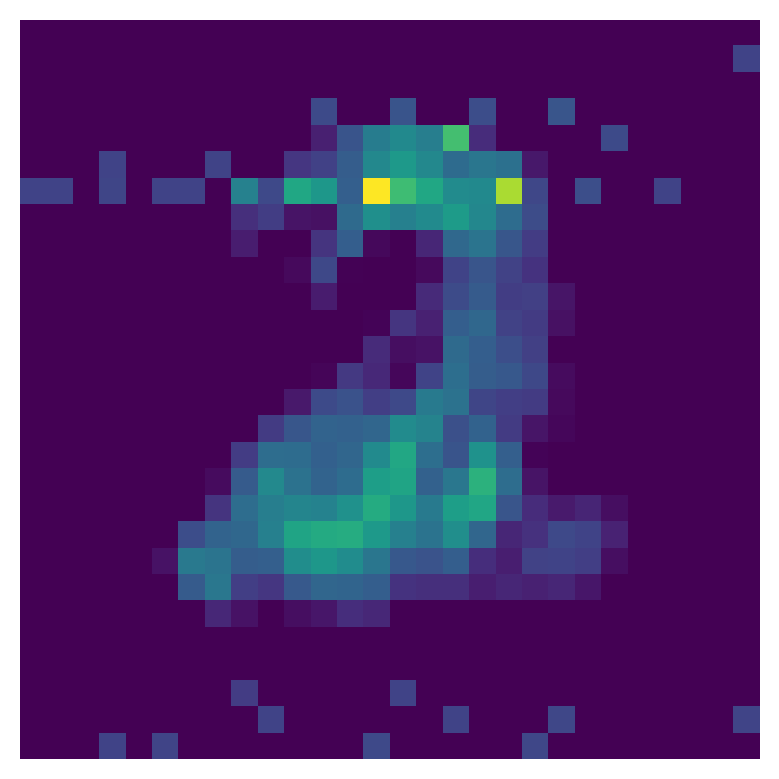} &
    \includegraphics[width=0.060\textwidth]{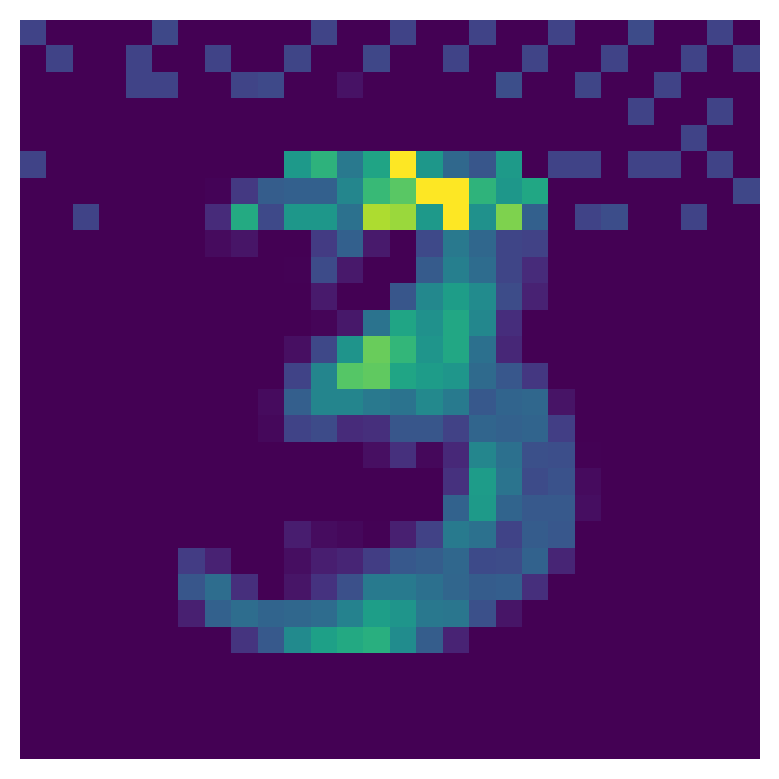} &
    \includegraphics[width=0.060\textwidth]{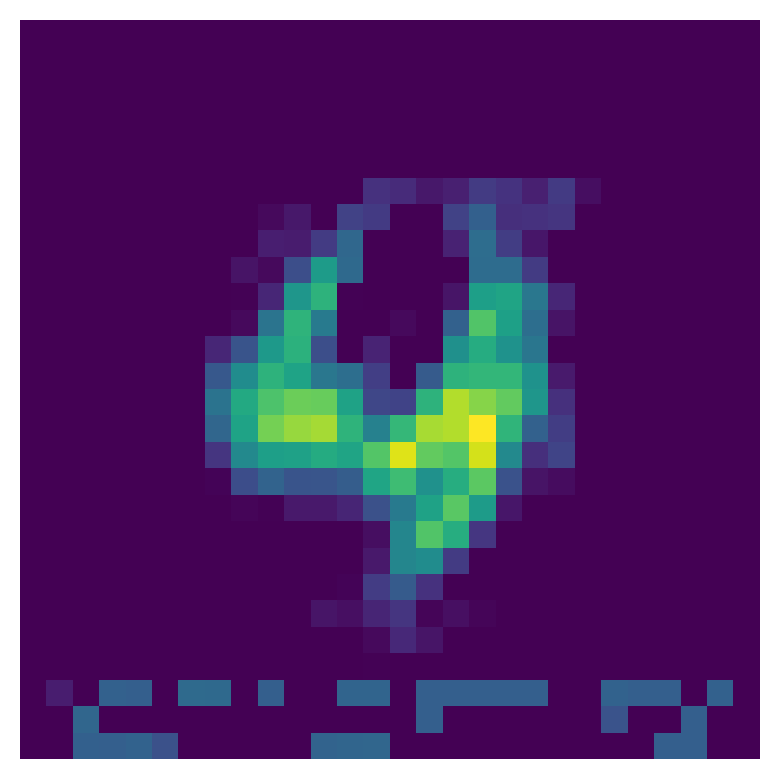} \\
     & \\
    \includegraphics[width=0.060\textwidth]{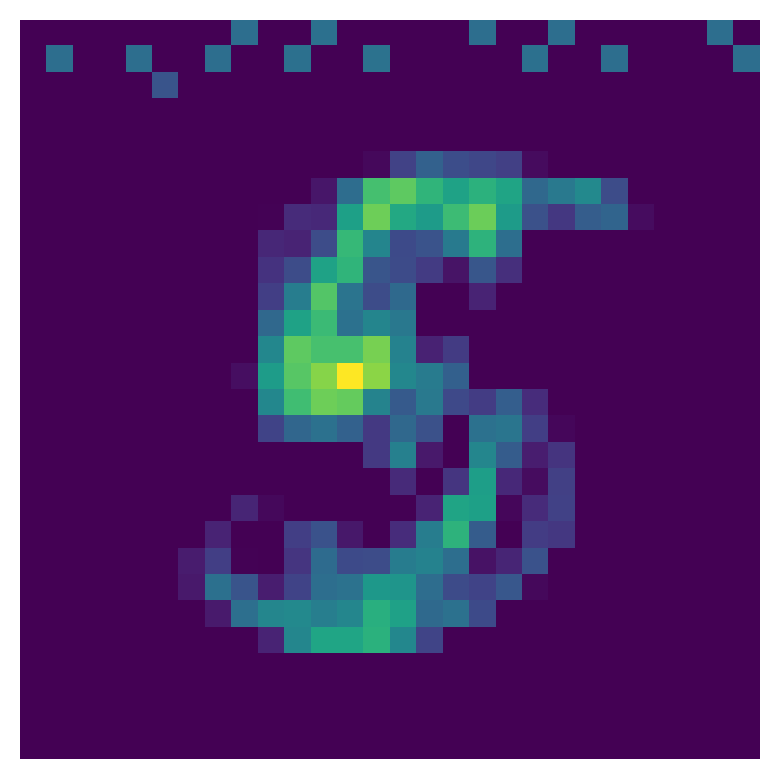} &
    \includegraphics[width=0.060\textwidth]{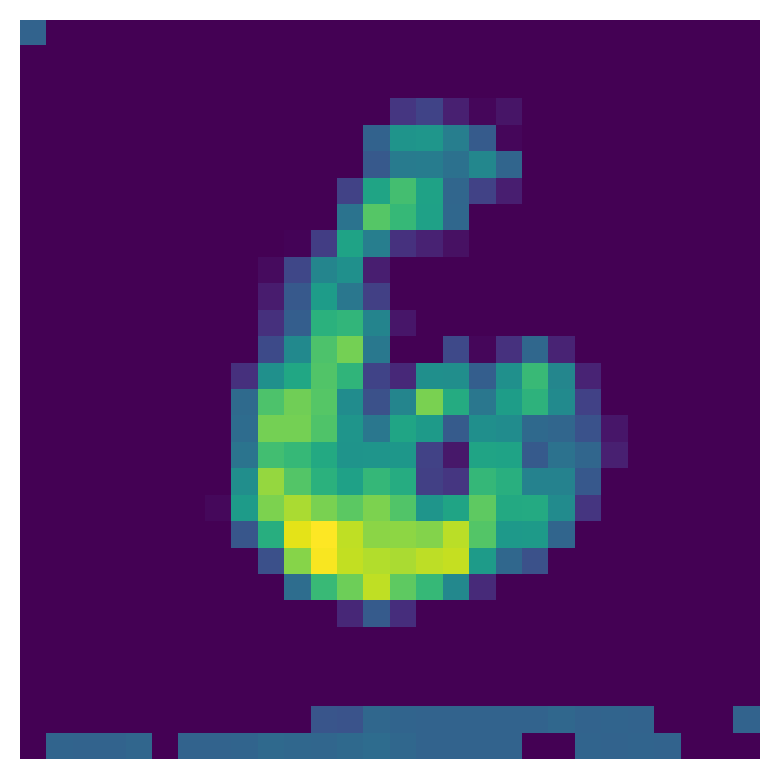} &
    \includegraphics[width=0.060\textwidth]{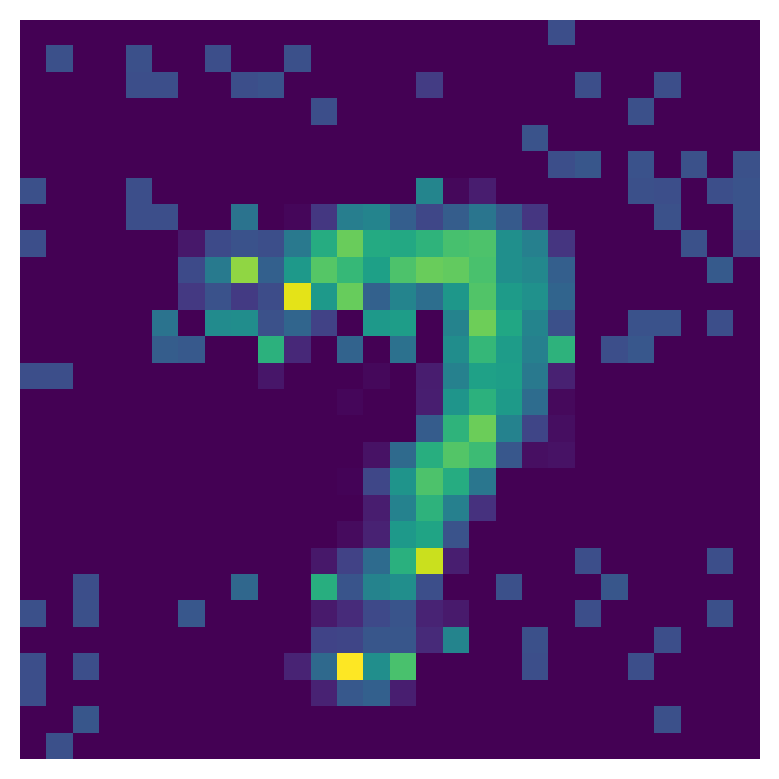} &
    \includegraphics[width=0.060\textwidth]{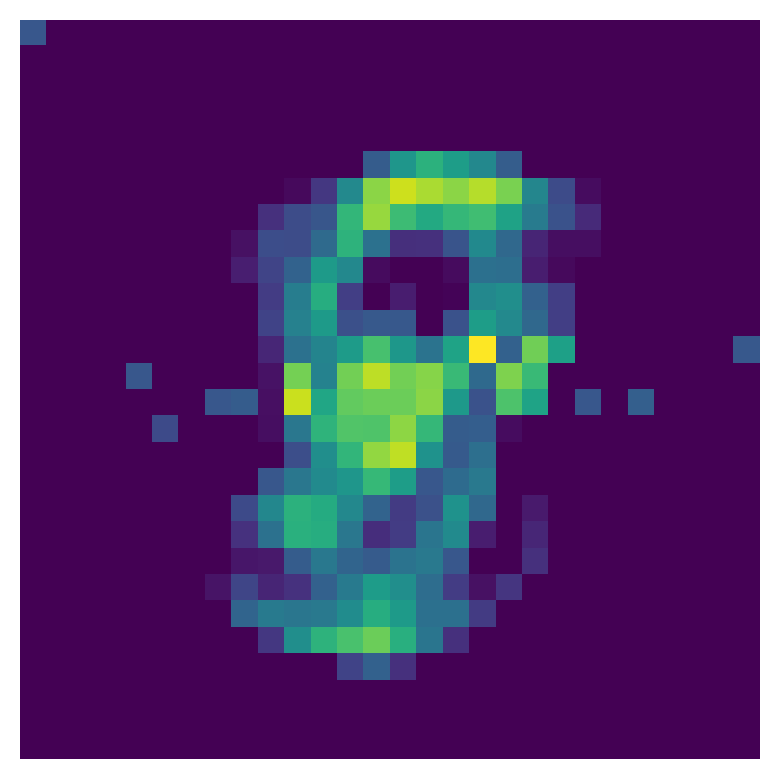} &
    \includegraphics[width=0.060\textwidth]{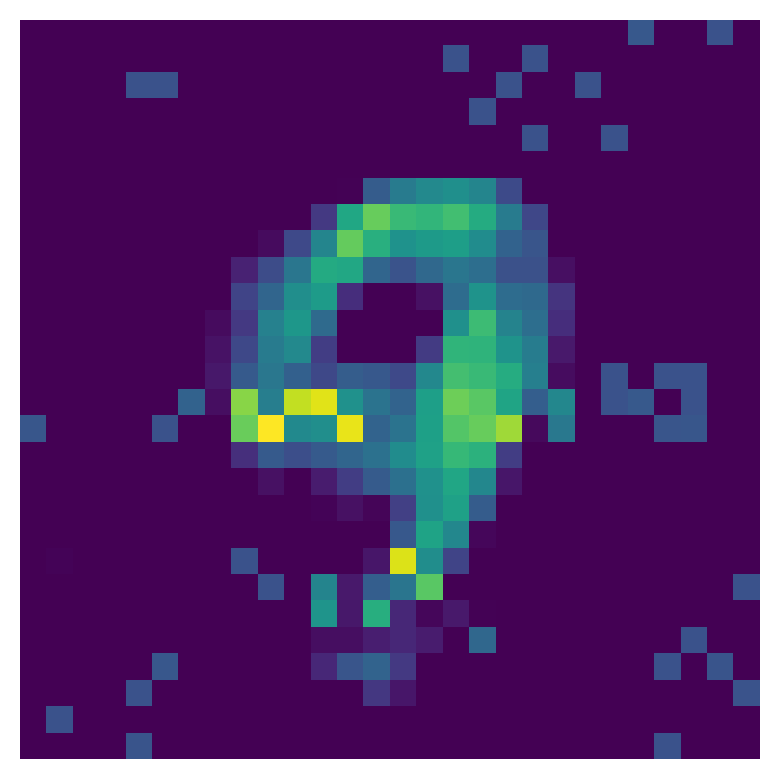} \\
    & \\
    \includegraphics[width=0.060\textwidth]{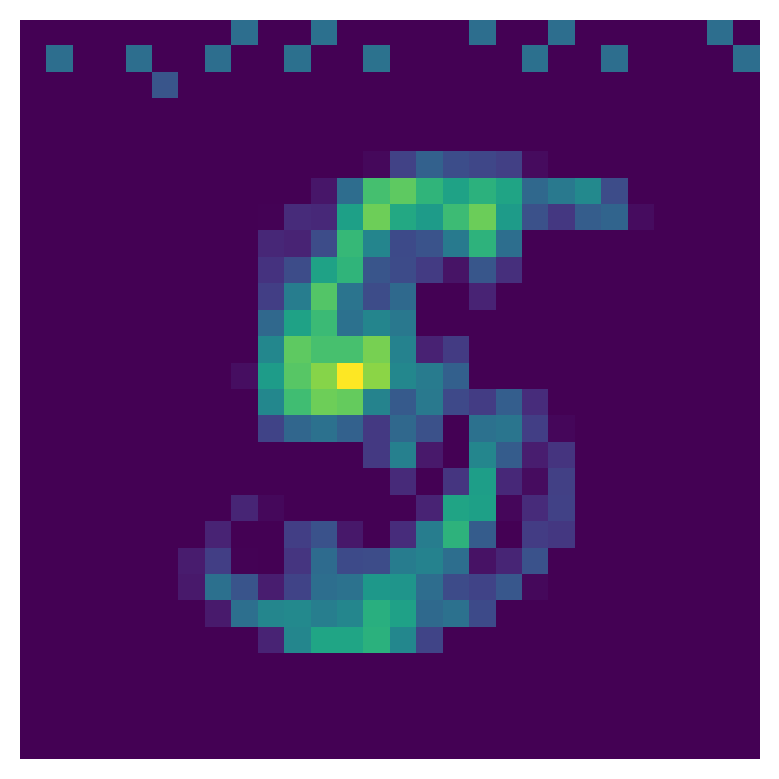} &
    \includegraphics[width=0.060\textwidth]{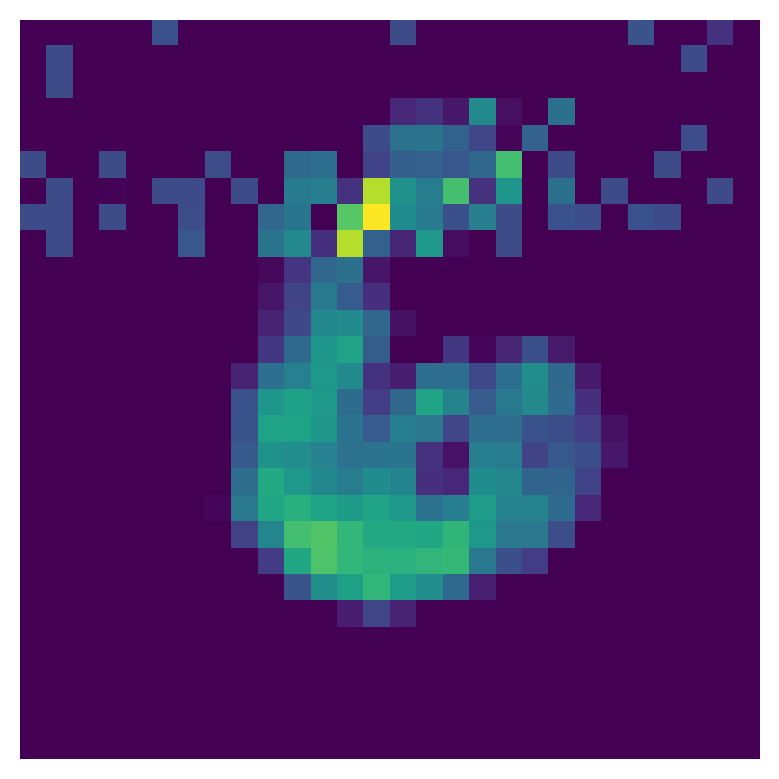} &
    \includegraphics[width=0.060\textwidth]{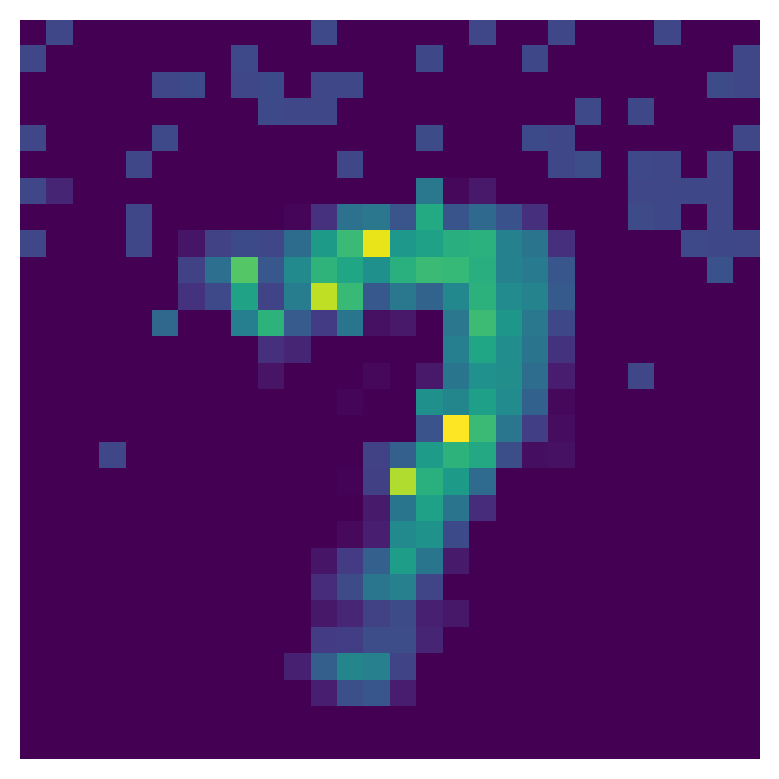} &
    \includegraphics[width=0.060\textwidth]{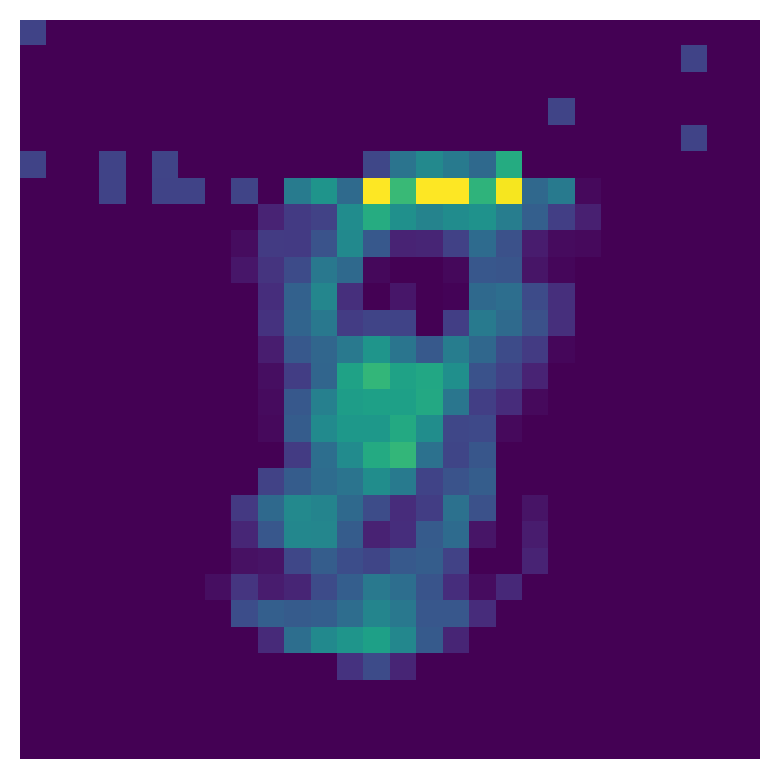} &
    \includegraphics[width=0.060\textwidth]{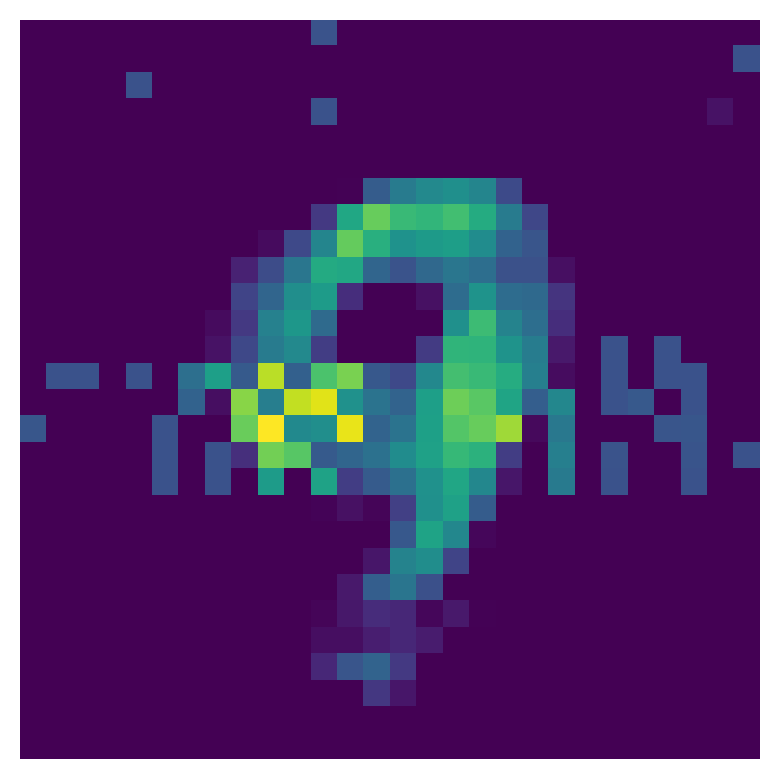} \\
    \end{tabular}
    \caption{Complete crafted examples for MNIST dataset. Line (1-2): Boundary and Adversarial Examples for class 0-4;
    Line (3-4): Boundary and Adversarial Examples for class 5-9.}
    \label{fig:MNIST}
\end{figure}

\begin{figure}[!htbp]
    \centering
    \tabcolsep=0.01\textwidth
    \begin{tabular}{ccccc}
    \includegraphics[width=0.060\textwidth]{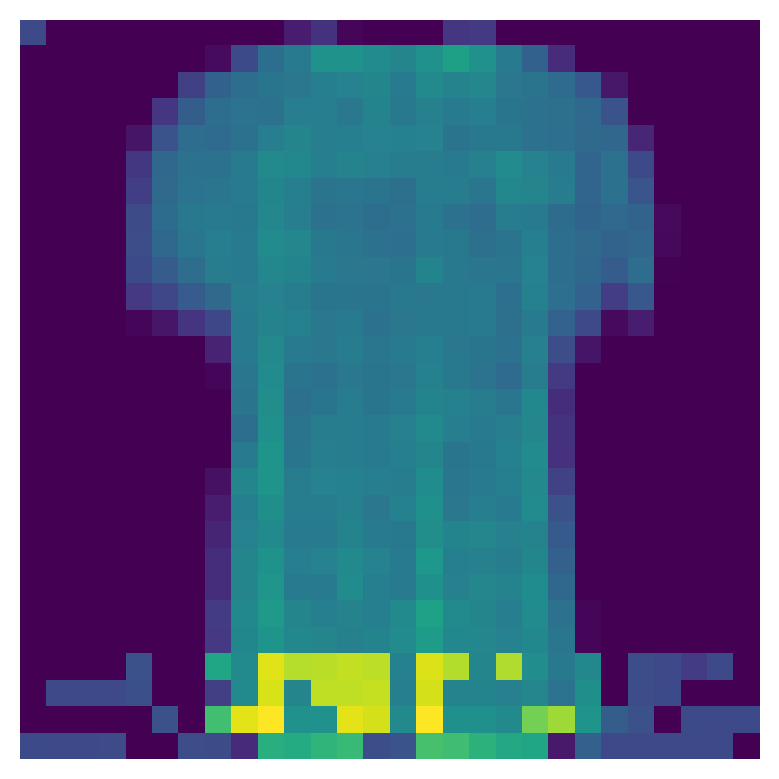} &
    \includegraphics[width=0.060\textwidth]{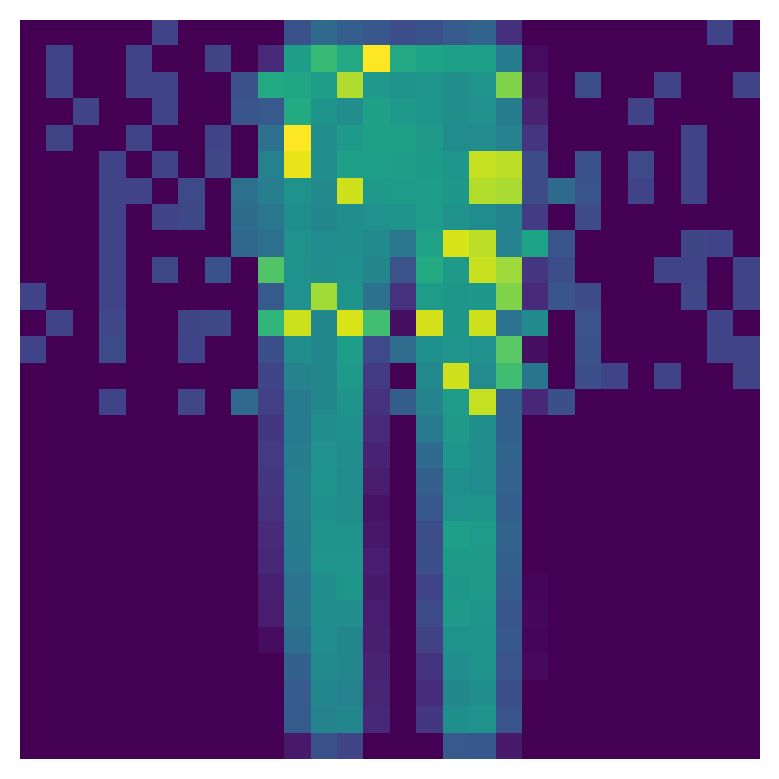} &
    \includegraphics[width=0.060\textwidth]{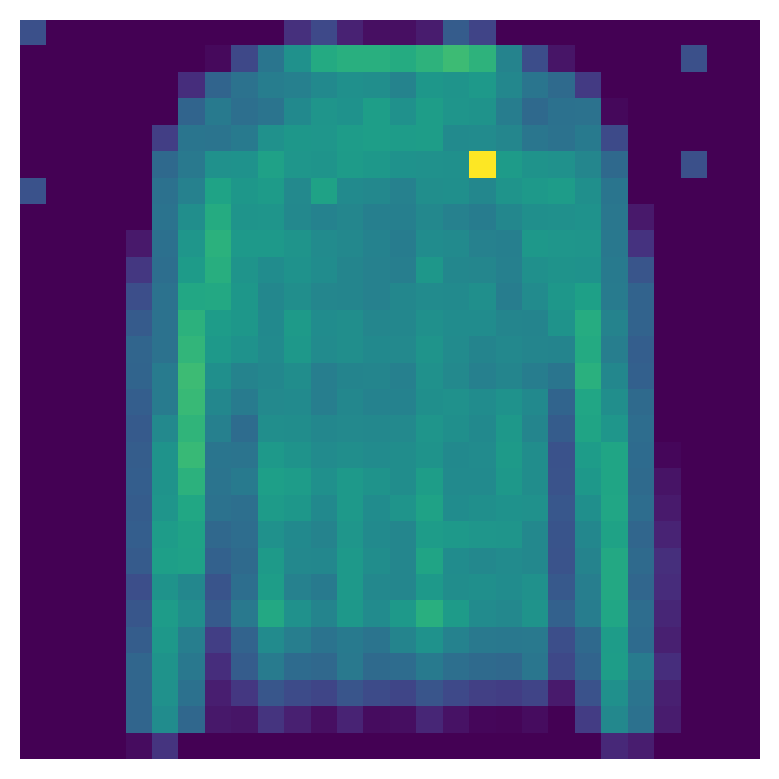} &
    \includegraphics[width=0.060\textwidth]{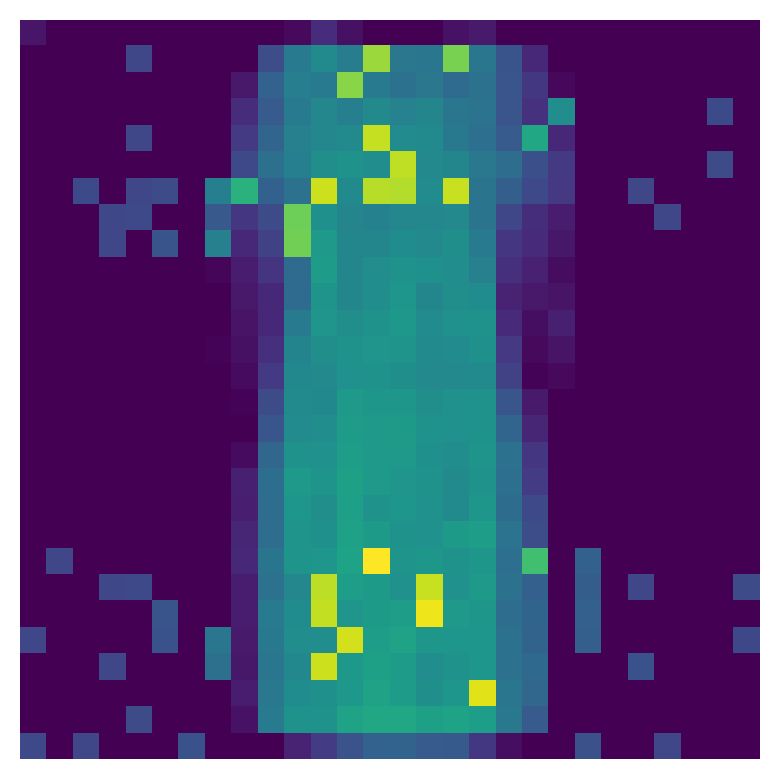} &
    \includegraphics[width=0.060\textwidth]{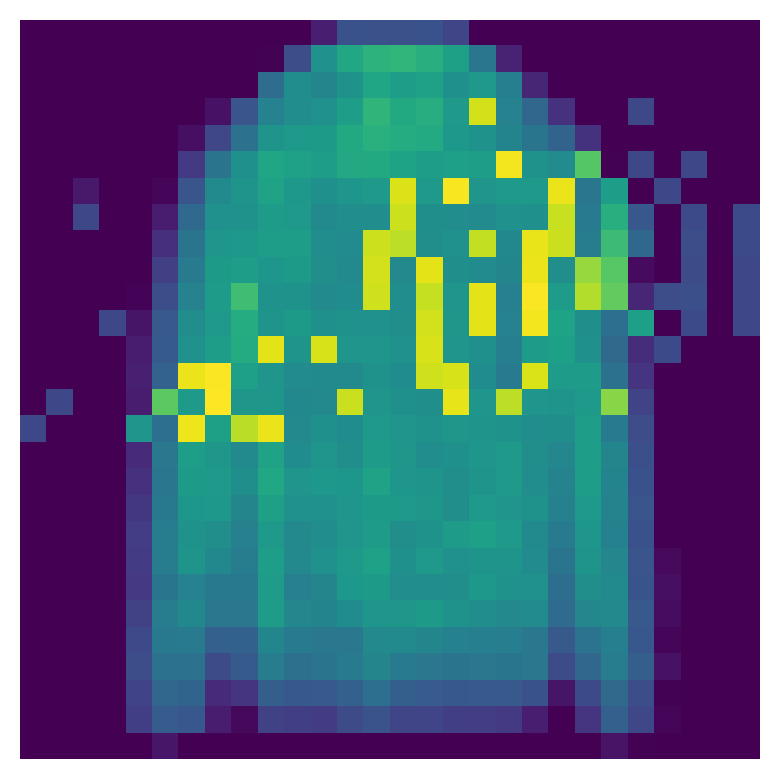} \\
    & \\
    \includegraphics[width=0.060\textwidth]{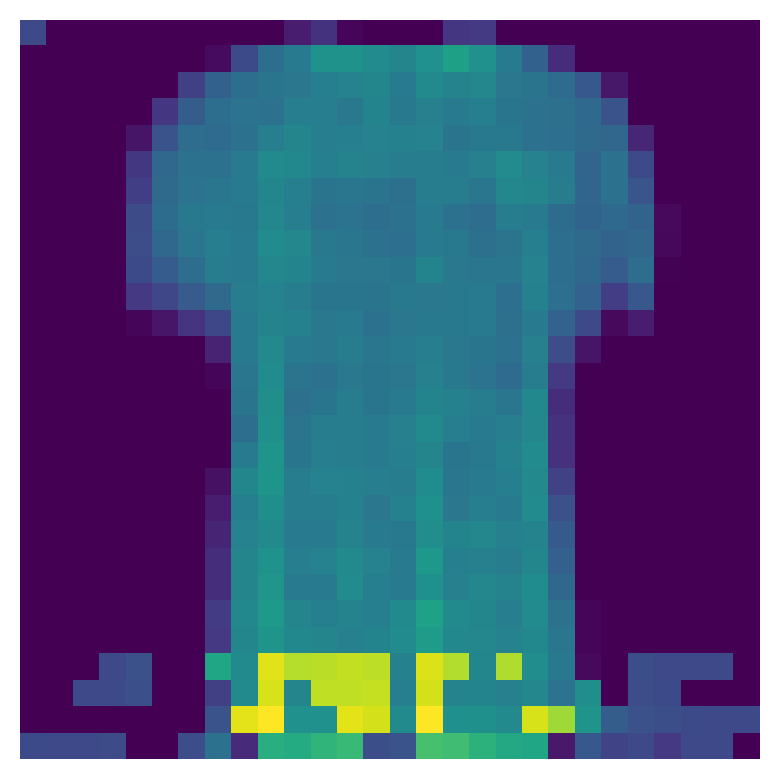} &
    \includegraphics[width=0.060\textwidth]{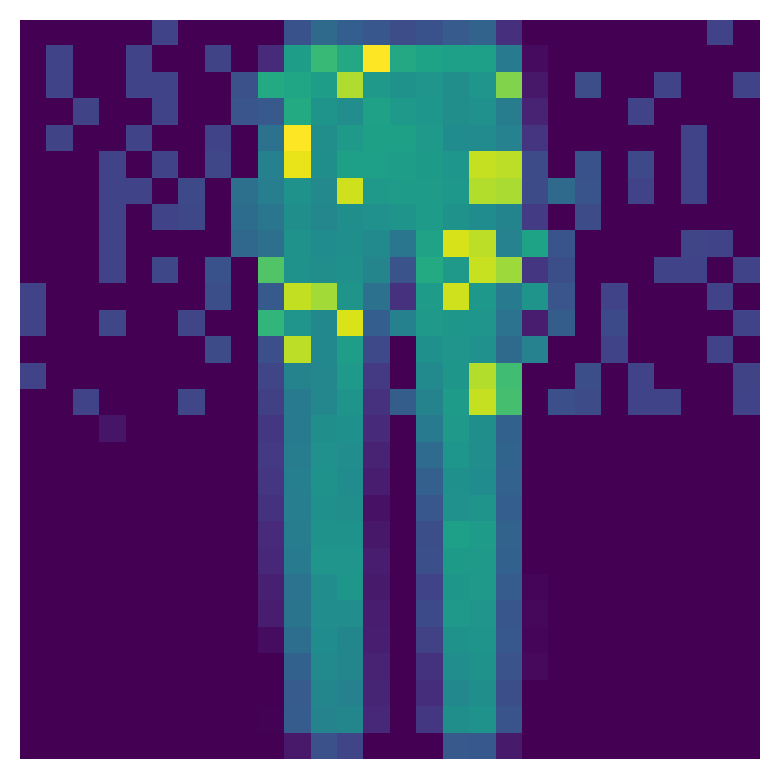} &
    \includegraphics[width=0.060\textwidth]{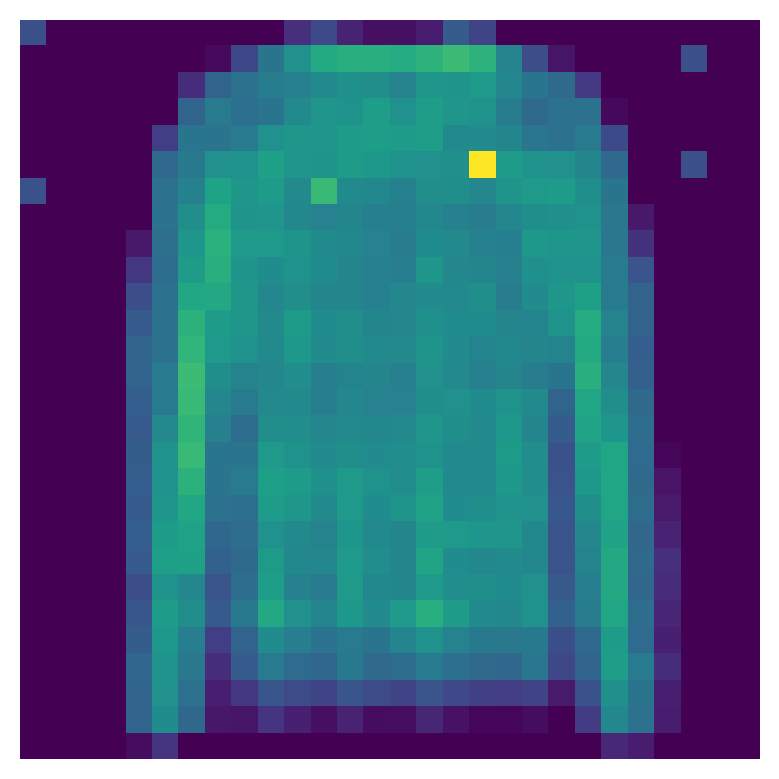} &
    \includegraphics[width=0.060\textwidth]{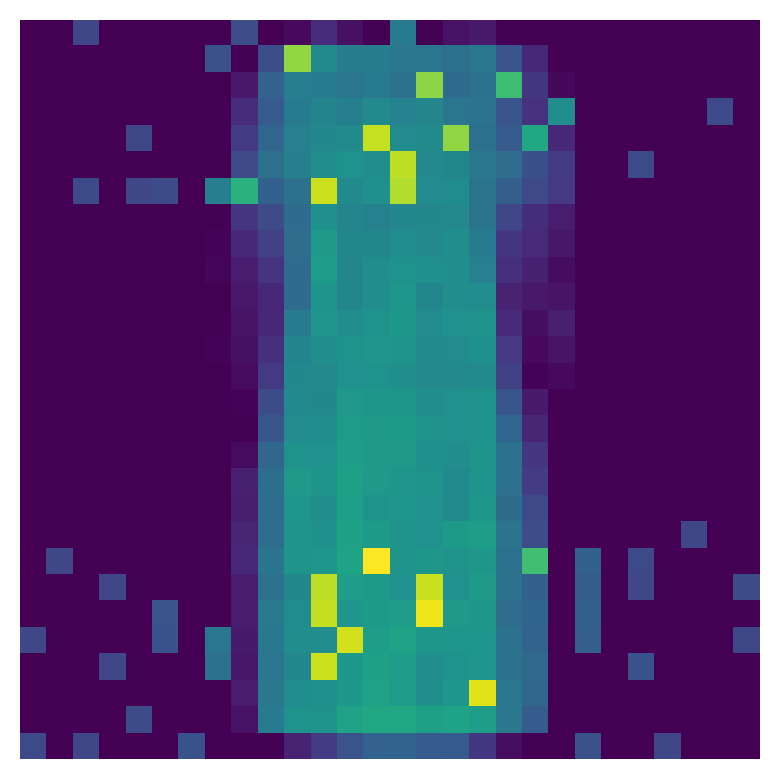} &
    \includegraphics[width=0.060\textwidth]{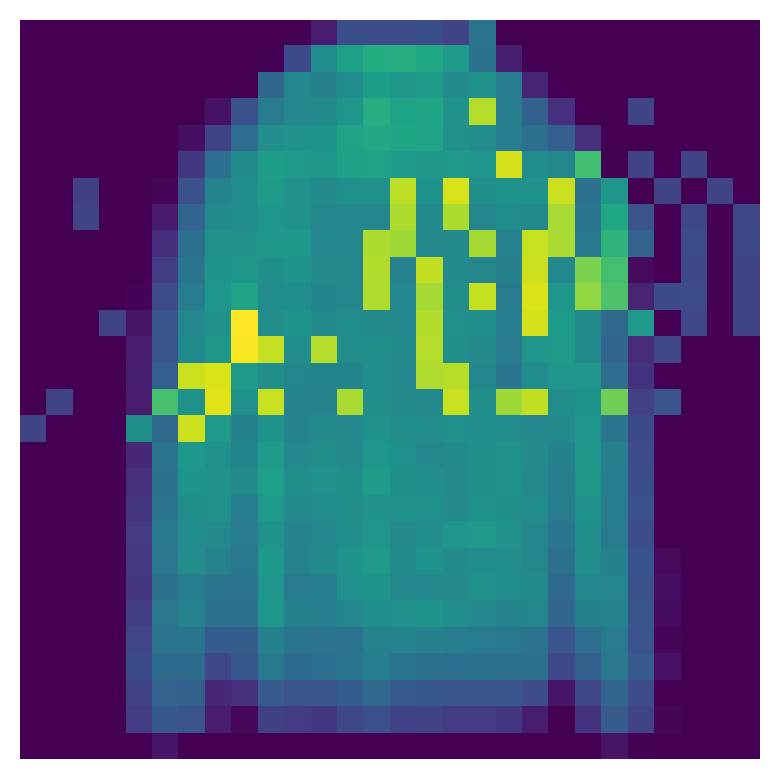} \\
     & \\
    \includegraphics[width=0.060\textwidth]{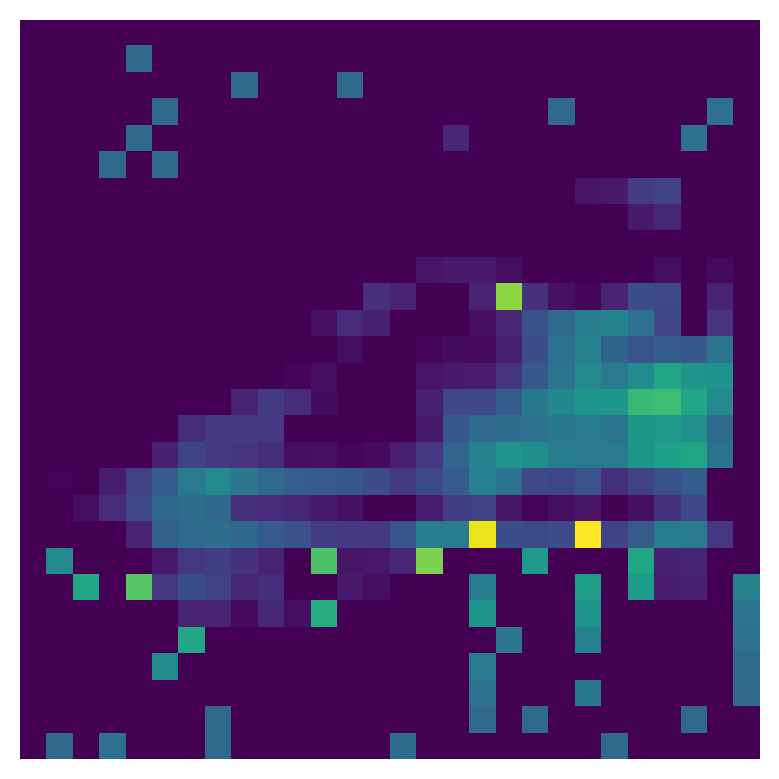} &
    \includegraphics[width=0.060\textwidth]{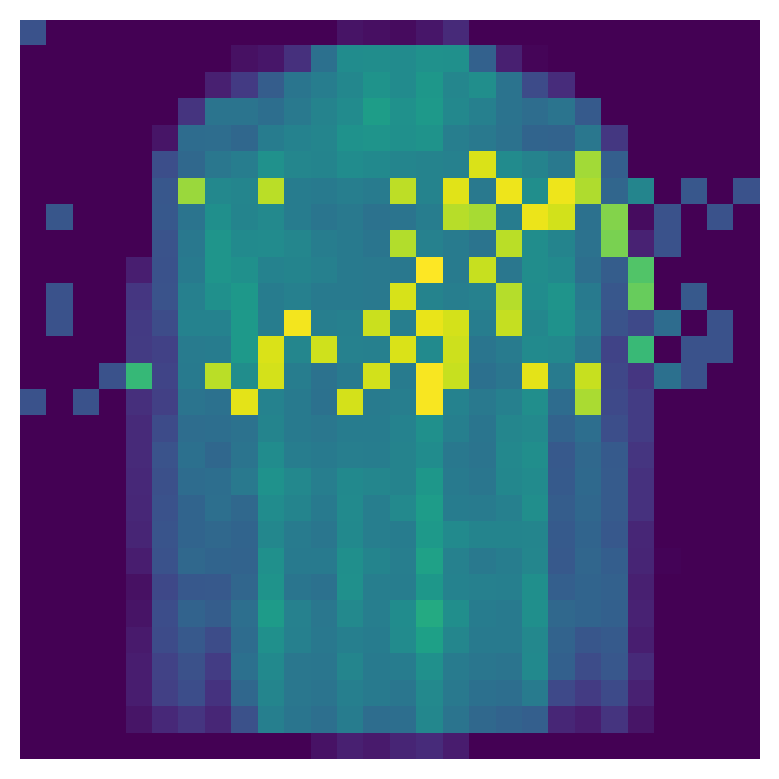} &
    \includegraphics[width=0.060\textwidth]{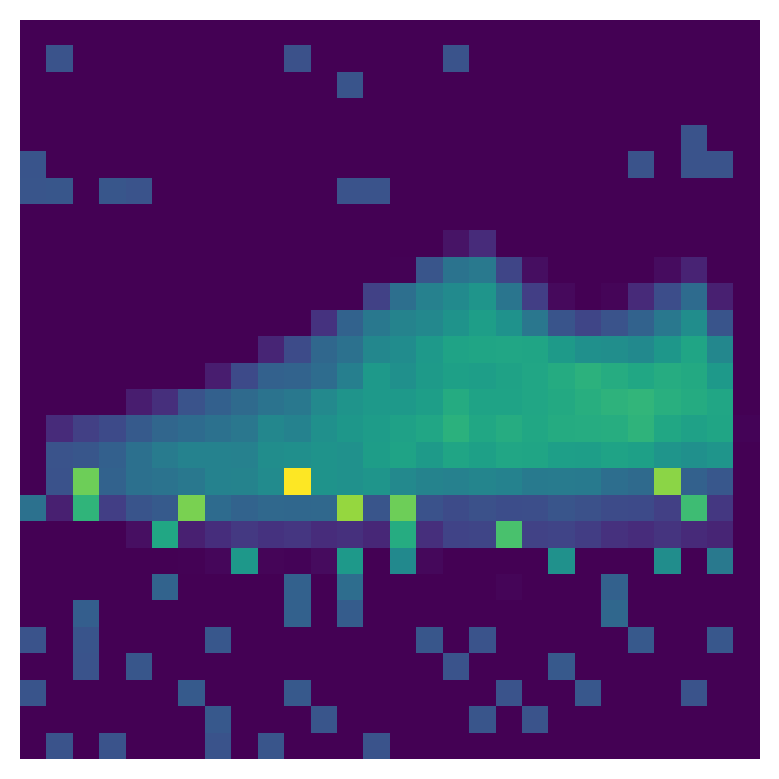} &
    \includegraphics[width=0.060\textwidth]{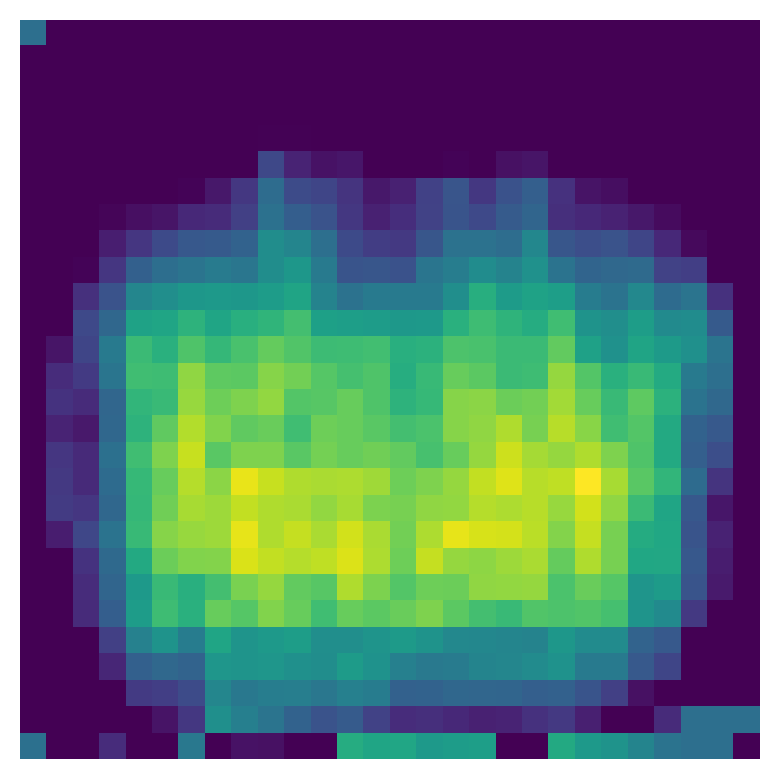} &
    \includegraphics[width=0.060\textwidth]{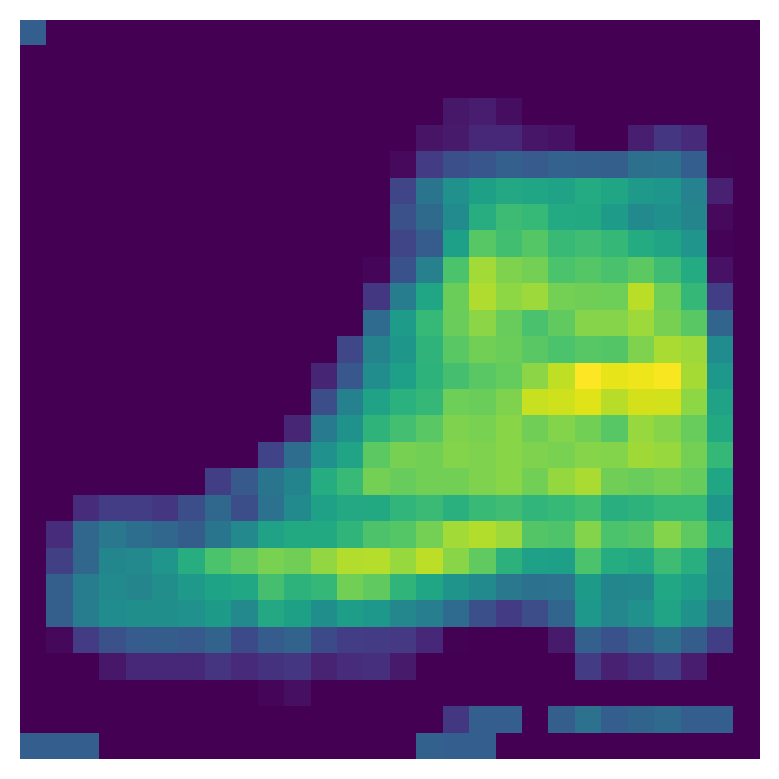} \\
    & \\
    \includegraphics[width=0.060\textwidth]{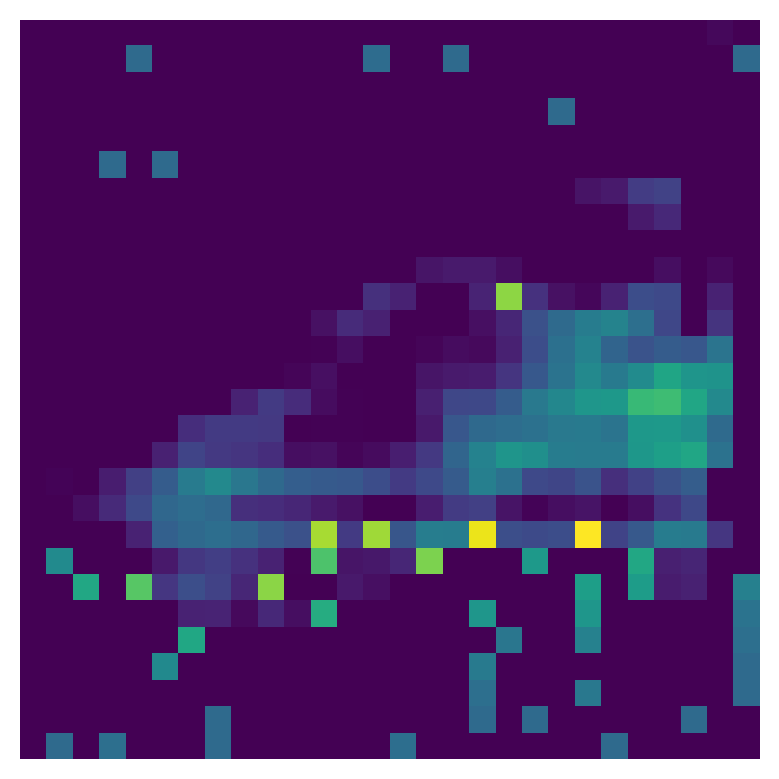} &
    \includegraphics[width=0.060\textwidth]{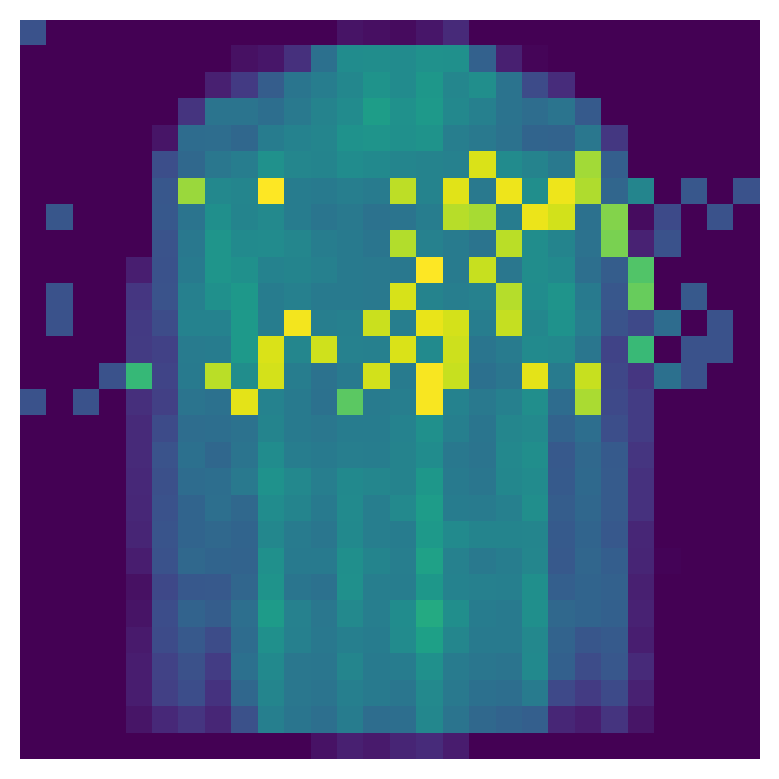} &
    \includegraphics[width=0.060\textwidth]{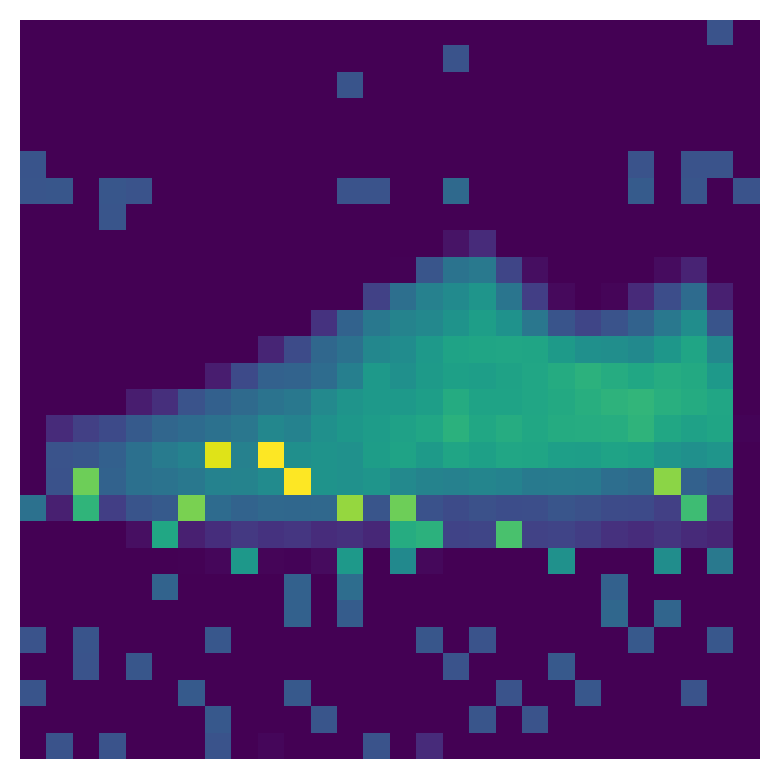} &
    \includegraphics[width=0.060\textwidth]{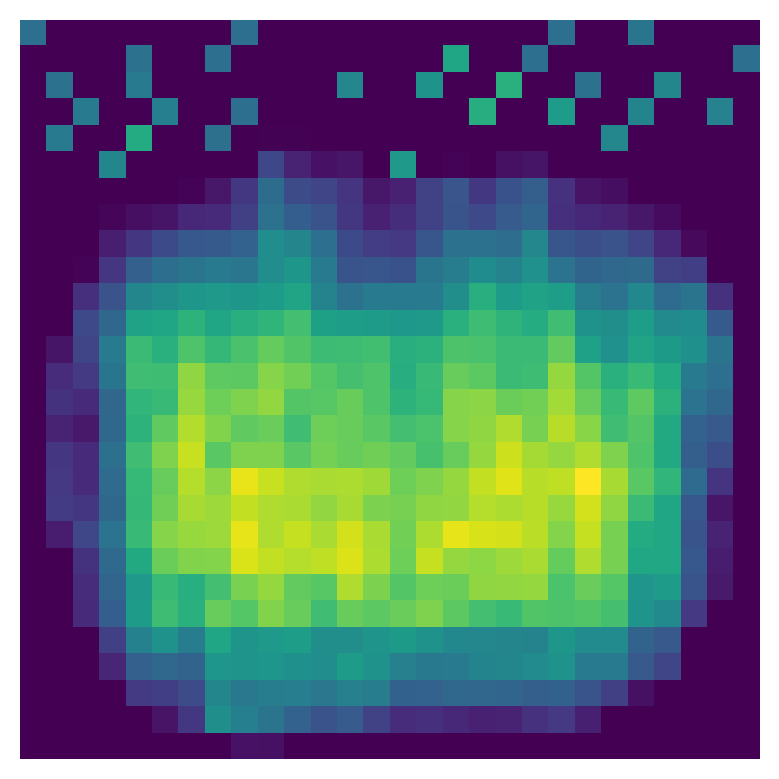} &
    \includegraphics[width=0.060\textwidth]{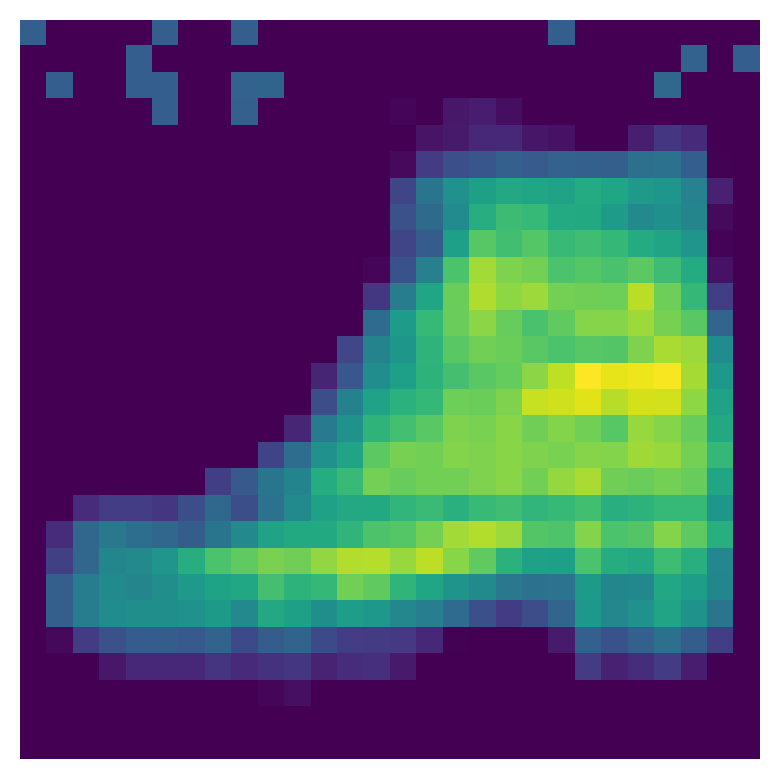} \\
    \end{tabular}
    \caption{Complete crafted examples for FashionMNIST dataset. Line (1-2): Boundary and Adversarial Examples for class 0-4;
    Line (3-4): Boundary and Adversarial Examples for class 5-9.}
    \label{fig:FashionMNIST}
\end{figure}

We first generate boundary samples for each class $i$. We identify samples $y$ that lie on the decision boundary between class $i$ and the adjacent class $j=i+1(\text{mod} 10)$. We set the radius of the adversarial region to 0.2. In cases where the expected samples are not generated within the time limit, we select another boundary class until success or all possibilities are exhausted.
A selection of boundary samples is shown in the first and third line of Figure~\ref{fig:MNIST} and~\ref{fig:FashionMNIST}.

Furthermore, by modifying the boundary constraint from $Output[i]==Output[j]$ to $Output[i] == Output[j] - \epsilon$, where $\epsilon>0$ is a small positive real number, we can use this framework to craft 
instances from the relaxed boundary regions.
The second line and fourth line of Figure~\ref{fig:MNIST} and~\ref{fig:FashionMNIST} show the crafted adversarial examples
from the relaxed regions.


\textbf{Towards Global Verification.} Finally, to identify all potential Adversarial Dangerous Regions (ADRs), one only needs to generate a sufficient number of boundary samples $y$ and use these samples to model the ADRs, as defined in Equation (\ref{eq:adr}). Note that the crafted boundary samples $y$ have already satisfied all
the conditions in the definition. 
Towards global robustness verification, we further propose an enhanced method to extract various prototypes for each class. Since samples with the same label can exhibit various patterns (such as 7 and \st{7} in MNIST), different prototypes in the same category can 
lead to different
ADRs. To address this issue, we utilize the K-means clustering algorithm~\cite{MacQueen1967SomeMF} to extract $K$ clusters from each class and calculate a prototype for each cluster. By doing so, we increase the granularity and globality of ADRs modeling.
See Figure~\ref{fig:100_pro} 
for an example.

\subsection{Related Work}
\label{related}
\subsubsection{Formal Modeling Machine Learning Models}
A representative approach to apply formal methods in modeling machine learning models is to extract a surrogate model for analysis and explaination, \textit{e.g.} decision graph for compiling Bayesian Networks~\cite{shih2018symbolic} and automata extraction for Recurrent Neural Networks~\cite{wei2022extracting}. The surrogate model paves way to model-based analysis with formal methods. 

Another thread of works attempt to leverage formal reasoning methods, such as provide explanations for decisions with abductive reasoning ~\cite{ignatiev2019abduction}. 
The tracability of formal explanations is also discussed in the literature~\cite{cooper2021tractability}.
Besides, a unifying logical framework for neural networks~\cite{zhang2022towards} is  proposed, which sheds light on the possibility of logical reasoning with network parameters.

\subsubsection{Formal Verification of FNNs}
Applying formal methods to verify FNNs~\cite{xiang2018verification} shows great superiority in safety-critical domains since they can provide provable guarantees. For the verification of adversarial robustness, so far, most existing works focus on the veification of local robustness which is based on  given test set~\cite{huang2017safety}.
Some improved techniques including
linear relaxation~\cite{zhang2022provably}, interval bound propagation (IBP)~\cite{zhang2019towards,wang2021beta} and reachability analysis~\cite{tran2019star} are successively proposed.
To the best of our knowledge, DeepGlobal~\cite{sun2021deepglobal,sun2022deepglobal} is currently the only one framework to verify the global robustness of FNNs, and we are the first to present complete specification and implementation of DeepGlobal with some certain SMT solver. 

\end{document}